\documentclass{article}

\usepackage[nonatbib,final]{nips_2016}

\usepackage{nips_2016}

\usepackage[dvips]{graphicx}\usepackage{amssymb,amsmath,color}

\usepackage[mathcal]{eucal}

\let\OLDthebibliography\thebibliography
\renewcommand\thebibliography[1]{
  \OLDthebibliography{#1}
  \setlength{\parskip}{2.4pt}
  \setlength{\itemsep}{0pt plus 0.3ex}
}

\usepackage{algorithmic,algorithm}

\usepackage{enumitem}
\setlist[itemize]{leftmargin=*}

\usepackage[utf8]{inputenc} 
\usepackage[T1]{fontenc}    
\usepackage{hyperref}       
\usepackage{url}            
\usepackage{booktabs}       
\usepackage{amsfonts}       
\usepackage{nicefrac}       
\usepackage{microtype}      

\usepackage{epstopdf}

\title{Stochastic Variance Reduction Methods \\
 for Saddle-Point Problems}

\author{
	P.~Balamurugan\\ 
	INRIA - Ecole Normale Sup\'erieure,
	Paris\\		
	\texttt{balamurugan.palaniappan@inria.fr}	
	 \And   
		Francis Bach \\
	INRIA - Ecole Normale Sup\'erieure,
	Paris\\
	\texttt{francis.bach@ens.fr}
}

\newcommand{\BEAS}{\begin{eqnarray*}}
\newcommand{\EEAS}{\end{eqnarray*}}
\newcommand{\BEA}{\begin{eqnarray}}
\newcommand{\EEA}{\end{eqnarray}}
\newcommand{\BEQ}{\begin{equation}}
\newcommand{\EEQ}{\end{equation}}
\newcommand{\BIT}{\begin{itemize}}
\newcommand{\EIT}{\end{itemize}}
\newcommand{\BNUM}{\begin{enumerate}}
\newcommand{\ENUM}{\end{enumerate}}
\newcommand{\BA}{\begin{array}}
\newcommand{\EA}{\end{array}}

\newcommand{\Diag}{\mathop{\rm Diag}}

\newcommand{\idm}{I}
\newcommand{\rb}{\mathbb{R}}
\newcommand{\BlackBox}{\rule{1.5ex}{1.5ex}}  

\newtheorem{theorem}{Theorem}

\newcommand{\mysec}[1]{Section~\ref{sec:#1}}
\newcommand{\eq}[1]{Eq.~(\ref{eq:#1})}

\def \E{{\mathbb E}}

 \def \DD{ {\small\big(\!\!\! \begin{array}{cc} 1/\lambda\! \!\!&\!\! 0 \\
 0 \!\!&\!\!\! 1/\gamma  \end{array}\!\!\big)}}
\def \DDtilde{ {\small\big(\!\!\! \begin{array}{cc} 1/\tilde\lambda\! \!\!&\!\! 0 \\
 0 \!\!&\!\!\! 1/\tilde\gamma  \end{array}\!\!\big)}}

\def \E{{\mathbb E}}
\def \I{{\mathcal I}}

\begin{document}

\maketitle

 \vspace*{-.111cm}

\begin{abstract}
We consider convex-concave saddle-point problems where the objective functions may be split in many components, and extend recent stochastic variance reduction methods (such as SVRG or SAGA) to provide the first  large-scale linearly convergent algorithms for this class of problems which are common in machine learning. While the algorithmic extension is straightforward, it comes with challenges and opportunities: (a) the convex minimization analysis does not apply and we use the notion of monotone operators to prove convergence, showing in particular that the same algorithm applies to a larger class of problems, such as variational inequalities,  (b) there are two notions of splits, in terms of functions, or in terms of partial derivatives, (c) the split does need to be done with convex-concave terms, (d) non-uniform sampling is key to an efficient algorithm, both in theory and practice, and (e)  these incremental algorithms can be easily accelerated using a simple extension of the ``catalyst'' framework,  leading to an algorithm which is always superior to accelerated batch algorithms.
  \end{abstract}

   \vspace*{-.111cm}

 \section{Introduction}
 
 \vspace*{-.111cm}
 
When using  optimization in machine learning, leveraging the natural separability of the objective functions has led to many algorithmic advances; the most common example is the separability as a sum of individual loss terms corresponding to individual observations, which leads to stochastic gradient descent techniques. Several lines of work have shown that the plain Robbins-Monro algorithm could be accelerated for strongly-convex finite sums, e.g., SAG~\cite{roux2012stochastic}, SVRG~\cite{svrg}, SAGA~\cite{defazio2014saga}. However, these only apply to separable objective functions.

In order to tackle non-separable losses or regularizers, we consider the  saddle-point problem:
 \BEQ
 \label{eq:saddle}
 \min_{x \in \rb^d} \max_{ y \in \rb^n} \ \ K(x,y) + M(x,y),
 \EEQ
where the functions $K $ and $M$ are ``convex-concave'', that is, convex with respect to the first variable, and concave with respect to the second variable, with $M$ potentially non-smooth but ``simple'' (e.g., for which the proximal operator is easy to compute), and $K$ smooth. These problems occur naturally within convex optimization through Lagrange or Fenchel duality~\cite{rockafellar1970monotone}; for example the bilinear saddle-point problem $\min_{x \in \rb^d} \max_{ y \in \rb^n} f(x) + y^\top K x - g(y)$ corresponds to a supervised learning problem with design matrix $K$, a loss function $g^\ast$  (the Fenchel conjugate of $g$) and a regularizer $f$.

We assume that the function $K$ may be split into a potentially large number of components. Many problems in machine learning exhibit that structure in the saddle-point formulation, but not in the associated convex minimization  and concave maximization problems (see examples in \mysec{examples}).

Like for convex minimization, gradient-based techniques that are blind to this separable structure   need to access all the components at every iteration. We show that algorithms such as SVRG~\cite{svrg} and SAGA~\cite{defazio2014saga} may be readily extended to the saddle-point problem. While the algorithmic extension is straightforward, it comes with challenges and opportunities. We make the following contributions:

 \vspace*{-.211cm}

\BIT
\item[--] We provide the first convergence analysis for these algorithms for saddle-point problems, which differs significantly from the associated convex minimization set-up. In particular, we use in \mysec{monotone} the interpretation of saddle-point problems as finding the zeros of a monotone operator, and only use the monotonicity properties to show linear convergence of our algorithms, thus showing that they extend beyond saddle-point problems, e.g., to variational inequalities~\cite{harker1990finite,zhu1996co}.
\item[--] We show that the saddle-point formulation  (a) allows two different notions of splits, in terms of functions, or in terms of partial derivatives, (b) does need splits into convex-concave terms (as opposed to convex minimization), and (c) that non-uniform sampling is key to an efficient algorithm, both in theory and practice (see experiments in \mysec{experiments}).

\item[--] We show in \mysec{acc} that these incremental algorithms can be easily accelerated using a simple extension of the ``catalyst'' framework of~\cite{lin2015universal}, thus leading to an algorithm which is always superior to accelerated batch algorithms.

\EIT

 \section{Composite Decomposable Saddle-Point Problems}
  \label{sec:comp}
  
 \vspace*{-.111cm}
 
We now present our new  algorithms on saddle-point problems and show a natural extension to monotone operators later in Section \ref{sec:monotone}.  
We thus consider the saddle-point problem defined in \eq{saddle}, with the following assumptions:
 
  \vspace*{-.211cm}

\BIT
 \item[(A)$\!$] $M$ is strongly $(\lambda,\gamma$)-convex-concave, that is, the function $ (x,y) \mapsto M(x,y) - \frac{\lambda}{2}\| x\|^2 + \frac{\gamma}{2} \| y\|^2$ is convex-concave. Moreover, we assume that we may compute the proximal operator of $M$, i.e., 
  for any $(x',y') \in \rb^{n +d}$ ($\sigma$ is the step-length parameter associated with the prox operator):
 
 \vspace*{-.25cm}

 \BEQ
 \label{eq:prox}
 {\rm prox}_{M}^\sigma(x',y') = \arg\min_{x \in \rb^d} \max_{ y \in \rb^n}  \ \  \sigma M(x,y)  +  \textstyle\frac{\lambda}{2}\| x - x'\|^2 - \frac{\gamma}{2} \| y - y'\|^2.
 \EEQ

 \vspace*{-.15cm}

 The values of $\lambda$ and $\gamma$ lead to the definition of a weighted Euclidean norm on $\rb^{n +d}$ defined as
  $\Omega(x,y)^2  = \lambda \| x \|^2 + \gamma \| y \|^2$, with dual norm defined through $\Omega^\ast(x,y)^2  = \lambda^{-1} \| x \|^2 + \gamma^{-1} \| y \|^2$. Dealing with the two different scaling factors $\lambda$ and $\gamma$ is crucial for good performance, as these may be very different, depending on the many arbitrary ways to set-up a saddle-point problem.
  
  \item[(B)$\!$] $K$ is convex-concave and has Lipschitz-continuous gradients; it is natural to consider the \emph{gradient operator} $B: \rb^{n + d} \to \rb^{n + d}$ defined as $B(x,y) = ( \partial_x K(x,y),  -\partial_y K(x,y) ) \in \rb^{n + d}$ and to consider $L = \sup_{\Omega(x-x',y-y') = 1}  {\Omega^\ast( B(x,y) - B(x',y') )}$. The quantity $L$ represents the condition number of the problem.

\item[(C)$\!$] The vector-valued function $B(x,y) = ( \partial_x K(x,y),  -\partial_y K(x,y) ) \in \rb^{n + d}$ may be split into a family of vector-valued functions as $B = \sum_{i \in \I} B_i$, where the only constraint is that each $B_i$ is Lipschitz-continuous (with constant $L_i$). There is no need to assume the existence of a function $K_i : \rb^{n + d} \to \rb$ such that $B_i = ( \partial_x K_i ,  -\partial_y K_i)$.

We will also consider splits which are adapted to the saddle-point nature of the problem, that is, of the form $B(x,y) = \big( \sum_{k \in \mathcal{K}}  B_{k}^x(x,y), \sum_{j \in \mathcal{J}} B_{j}^y(x,y) \big)$, which is a subcase of the above with $\I = \mathcal{J} \times \mathcal{K}$, $B_{jk}(x,y) = ( p_j  B_{k}^x(x,y), q_k B_j^y(x,y) )$, for $p$ and $q$ sequences that sum to one. This substructure, which we refer to as ``factored'', will only make a difference when storing the values of these operators in \mysec{saga} for  our SAGA algorithm.

 \EIT
 
 \vspace*{-.1cm}

 Given assumptions (A)-(B), the saddle-point problem in \eq{saddle} has a unique solution $(x_\ast,y_\ast)$ such that
 $K(x_\ast,y) \!+ \! M(x_\ast,y) \leqslant K(x_\ast,y_\ast) \!+\! M(x_\ast,y_\ast)  \leqslant K(x,y_\ast) \! +\! M(x,y_\ast)$, for all $(x,y)$; moreover
$ \min_{x \in \rb^d} \max_{ y \in \rb^n}  \ K(x,y) + M(x,y) = \max_{ y \in \rb^n} \min_{x \in \rb^d}   \ K(x,y) + M(x,y)$~(see, e.g.,~\cite{bauschke2011convex,rockafellar1970monotone}).

The main generic examples for the functions $K(x,y)$ and $M(x,y)$ are:

  \vspace*{-.211cm}

\BIT
\item[--] \textbf{Bilinear saddle-point problems}: $K(x,y) = y^\top K x$ for a matrix $K \in \rb^{n \times d}$ (we identify here a matrix with the associated bilinear function), for which the vector-valued function $B(x,y)$ is linear, i.e., $B(x,y) = (K^\top y , - K x)$. Then $L =  \| K\|_{\rm op}  / \sqrt{\gamma \lambda}$, where $\| K\|_{\rm op}$ is the largest singular value of $K$.

There are two natural potential splits with $\I = \{1,\dots,n\} \times \{1,\dots,d\}$, with $B  = \sum_{j=1}^n \sum_{k=1}^d B_{jk}$: (a) the split into individual elements
  $B_{jk}(x,y) = K_{jk} (y_j , - x_k )$, where every element is the gradient operator of a bi-linear function, and (b) the ``factored'' split into rows/columns   $B_{jk}(x,y) = (q_k y_j K_{j\cdot}^\top , - p_j x_k K_{\cdot k})$, where 
  $K_{j \cdot }$ and $K_{\cdot k}$ are the $j$-th row and $k$-th column of $K$,  
  $p$ and $q$ are any set of vectors summing to one, and every element is not the gradient operator of any function. 
These splits correspond to several ``sketches'' of the matrix~$K$~\cite{woodruff2014sketching}, adapted to subsampling of $K$, but other sketches could be considered.

\item[--] \textbf{Separable functions}: $M(x,y) = f(x) - g(y)$ where $f$ is any $\lambda$-strongly-convex and $g$ is $\gamma$-strongly convex, for which the proximal operators  
${\rm prox}_{f}^\sigma(x') = \arg\min_{x \in \rb^d}   \  \sigma f(x)  + \frac{\lambda}{2}\| x - x'\|^2 $ and
${\rm prox}_{g}^\sigma(y') = \arg\max_{y\in \rb^d}   \  - \sigma g(y)   - \frac{\gamma}{2}\| y - y'\|^2 $  are easy to compute. In this situation,  ${\rm prox}_{M}^\sigma(x',y') = ( {\rm prox}_{f}^\sigma(x') , {\rm prox}_{g}^\sigma(y')  )$.
Following the usual set-up of composite optimization~\cite{Beck2009}, no smoothness assumption is made on $M$ and hence on $f$ or $g$.
\EIT

 \subsection{Examples in machine learning}
 
\vspace*{-.051cm}

\label{sec:examples}
Many  learning problems are formulated as convex optimization problems, and hence by duality as saddle-point problems. 
We now give examples where our new algorithms are particularly adapted.

\textbf{Supervised learning with non-separable losses or regularizers.}
For regularized linear supervised learning, with $n$ $d$-dimensional observations put in a design matrix $K \in \rb^{n \times d}$, the predictions are parameterized by a vector $x \in \rb^d$ and lead to a vector of predictions $K x \in \rb^n$. Given a loss function defined through its Fenchel conjugate $g^\ast$ from $\rb^n$ to $\rb$, and a regularizer $f(x)$, we obtain exactly a bi-linear saddle-point problem. When the loss $g^\ast$ or the regularizer $f$ is separable, i.e., a sum of functions of individual variables, we may apply existing fast gradient-techniques~\cite{roux2012stochastic,svrg,defazio2014saga} to the primal problem $\min_{x \in \rb^d} g^\ast(Kx) + f(x)$ or the dual problem
$\max_{y \in \rb^n} - g(y) - f^\ast(K^\top y)$, as well as methods dedicated to separable saddle-point problems~\cite{zhu2015adaptive,zhang2015stochastic}. When the loss $g^\ast$ and the regularizer~$f$ are not separable (but have a simple proximal operator), our new fast algorithms are the only ones that can be applied from the class of large-scale linearly convergent algorithms.

Non-separable \emph{losses} may occur when (a) predicting by affine functions of the inputs and not penalizing the constant terms (in this case defining the loss functions as the minimum over the constant term, which becomes non-separable) or (b) using structured output prediction methods that lead to convex surrogates to the area under the ROC curve (AUC) or other precision/recall quantities~\cite{joachims2005support,herbrich1999large}. These come often with efficient proximal operators (see \mysec{experiments} for an example).

Non-separable \emph{regularizers} with available efficient proximal operators are numerous, such as grouped-norms with potentially overlapping groups, norms based on submodular functions, or total variation (see~\cite{bach2012optimization} and references therein, and an example in \mysec{experiments}).

\textbf{Robust optimization.} The framework of robust optimization~\cite{ben2009robust} aims at optimizing an objective function with uncertain data. Given that the aim is then to minimize the maximal value of the objective function given the uncertainty, this leads naturally to saddle-point problems. 

\textbf{Convex relaxation of unsupervised learning.} Unsupervised learning leads to convex relaxations  which often exhibit structures naturally amenable to saddle-point problems, e.g, for discriminative clustering~\cite{xu2004maximum} or matrix factorization~\cite{bach2008convex}.

\subsection{Existing batch algorithms}
\label{sec:batch}

\vspace*{-.051cm}

In this section, we review existing algorithms aimed at solving the composite saddle-point problem in \eq{saddle}, without using the sum-structure.  Note that it is often possible to apply batch algorithms for the associated primal or dual problems (which are not separable in general).

\textbf{Forward-backward (FB) algorithm.} The main iteration is 

\vspace*{-.75cm}

\BEAS
(x_t,y_t) & = &  {\rm prox}_{M}^\sigma \big[ (x_{t-1},y_{t-1}) - \sigma \DD B(x_{t-1},y_{t-1} ) \big]
\\[-.1cm]
& =  &  {\rm prox}_{M}^\sigma \big( x_{t-1} - \sigma \lambda^{-1} \partial_x K(x_{t-1},y_{t-1} ) + \sigma\gamma^{-1} \partial_y K(x_{t-1},y_{t-1} ))
.
\EEAS

\vspace*{-.25cm}

The algorithm aims at simultaneously minimizing with respect to $x$ while maximizing with respect to $y$ (when $M(x,y)$ is the sum of isotropic quadratic terms and indicator functions, we get simultaneous projected gradient descents).
This algorithm is known not to converge in general~\cite{bauschke2011convex}, but is linearly convergent for \emph{strongly}-convex-concave problems, when $\sigma = 1/L^2$, with the rate $( 1 - \textstyle 1/({1+L^2}) )^t$~\cite{chen1997convergence} (see simple proof in Appendix~\ref{app:fb}). This is the one we are going to adapt to stochastic variance reduction.

When $M(x,y) = f(x) - g(y)$, we obtain the two parallel updates $x_{t} = {\rm prox}_{f}^\sigma \big( x_{t-1} - \lambda^{-1}\sigma \partial_x K(x_{t-1},y_{t-1} \big) \big)$
and $y_{t} = {\rm prox}_{g}^\sigma \big( y_{t-1} + \gamma^{-1}\sigma \partial_y K(x_{t-1},y_{t-1} \big)\big)$, which can de done serially by replacing the second one by $y_{t} = {\rm prox}_{g}^\sigma \big( y_{t-1} + \gamma^{-1}\sigma \partial_y K(x_{t},y_{t-1} \big) \big)$. This is often referred to as the Arrow-Hurwicz method (see~\cite{chambolle2011first} and references therein).


%

\textbf{Accelerated forward-backward algorithm.}
The forward-backward algorithm may be accelerated by a simple extrapolation step, similar to Nesterov's acceleration for convex minimization~\cite{nest2004}. The algorithm from~\cite{chambolle2011first}, which \emph{only applies to bilinear functions $K$}, and  which we extend from separable $M$ to our more general set-up in Appendix~\ref{app:acc}, has the following iteration:
\[
(x_t,y_t ) =  {\rm prox}_{M}^\sigma \big[ (x_{t-1},y_{t-1}) - \sigma \DD B(x_{t-1} + \theta ( x_{t-1} - x_{t-2} ),y_{t-1}
+ \theta ( y_{t-1} - y_{t-2} ) ) \big].
\]
With $\sigma =  {1}/{(2L)}$ and $\theta =  {L}/{(L+1)}$, we get an improved convergence rate, where $  ( 1 - 1/({1+L^2}) )^t$ is replaced by
$  ( 1 -1/({1+2 L}) )^t$. This is always a strong improvement when $L$ is large (ill-conditioned problems), as illustrated in \mysec{experiments}. Note that our acceleration technique in \mysec{acc} may be extended to get a similar rate for the batch set-up for non-linear $K$.

\subsection{Existing stochastic algorithms}
\label{sec:sto}

\vspace*{-.051cm}

 Forward-backward algorithms have been studied with added noise~\cite{rosasco2014stochastic}, leading to a convergence rate in $O(1/t)$ after $t$ iterations for strongly-convex-concave problems. In our setting, we replace $B(x,y)$ in our algorithm with $\frac{1}{\pi_i} B_i (x,y)$, where $i \in \I$ is sampled from the probability vector $(\pi_i)_i$ (good probability vectors will depend on the application, see below for bilinear problems). We have $\E B_i(x,y) = B(x,y)$; the main iteration is then
 
 \vspace*{-.5cm}

\[ \textstyle (x_t,y_t) =   {\rm prox}_{M}^{\sigma_t} \big[ (x_{t-1},y_{t-1}) - \sigma_t \DD \frac{1}{\pi_{i_t}}B_{i_t} (x_{t-1},y_{t-1} ) \big] , \]

 \vspace*{-.25cm}
 
 with  $i_t$ selected  independently at random in $\I$ with probability vector $\pi$. In Appendix~\ref{app:sto}, we show that using $\sigma_t =2/ ( t + 1 +  8 \bar{L}(\pi)^2 )$ leads to a convergence rate in $O(1/t)$, where $\bar{L}(\pi)$ is a smoothness constant explicited below. For saddle-point problems, it leads to the complexities shown in Table~\ref{table:summary}. Like for convex minimization, it is fast early on but the performance levels off. Such schemes are typically used in sublinear algorithms~\cite{clarkson2012sublinear}.

\subsection{Sampling probabilities, convergence rates and running-time complexities}

\vspace*{-.051cm}

In order to characterize running-times, we denote by $T(A)$ the complexity of computing $A (x,y)$ for any operator $A$ and $(x,y) \in \rb^{n + d}$, while we denote by $T_{\rm prox}(M)$ the complexity of computing ${\rm prox}_{M}^\sigma(x,y)$.  
In our motivating example of bilinear functions $K(x,y)$, we assume that $T_{\rm prox}(M)$ takes times proportional to $n+d$ and getting a single element of $K$ is $O(1)$.

In order to characterize the convergence rate, we need the Lipschitz-constant $L$ (which happens to be the condition number with our normalization) defined earlier as well as a smoothness constant adapted to our sampling schemes:
\BEAS
\bar{L}(\pi)^2 & = &  \textstyle\sup_{(x,y,x',y')}    \sum_{i \in \I} \frac{1}{\pi_i} \Omega^\ast( B_i (x,y)  - B_i (x',y'))^2 \mbox{ such that }\Omega(x-x',y-y')^2 \leqslant 1.
\EEAS
We always have the bounds $ L^2 \leqslant \bar{L}(\pi)^2 \leqslant \max_{i \in \I} L_i^2 \times \sum_{i\in \I} \frac{1}{\pi_i}$. However, in structured situations (like in  bilinear saddle-point problems), we get  much improved bounds, as described below.

\textbf{Bi-linear saddle-point.}
The constant $L$ is equal to $\| K\|_{\rm op}/\sqrt{\lambda\gamma}$, and we will consider as well the Frobenius norm $\|K\|_F$ defined through $\|K\|_F^2 = \sum_{j,k} K_{jk}^2$, and the norm $\| K \|_{\max}$ defined as $\| K \|_{\max}^2 = \max\{  
\sup_j  (KK^\top)_{jj}^{1/2} ,    \sup_k (K^\top K)_{kk}^{1/2}
 \}$. Among the norms above, we always have:
\BEQ
\label{eq:norms}
\|K\|_{\rm max} \leqslant \| K\|_{\rm op} \leqslant \| K\|_F \leqslant \sqrt{ \max\{n,d\}} \| K \|_{\rm max} 
\leqslant \sqrt{ \max\{n,d\}} \| K \|_{\rm op} ,  
\EEQ
which allows to show below that some algorithms have better bounds than others. 

There are several schemes to choose the probabilities $\pi_{jk}$ (individual splits) and $\pi_{jk} = p_j q_k$ (factored splits). For the factored formulation where we select random rows and columns, we consider the non-uniform schemes $p_j =  { (KK^\top)_{jj} }/{\| K \|_F^2}$ and $q_k =  {  (K^\top K )_{kk} }/{ \| K \|_F^2}$, leading to $\bar{L}(\pi) \leqslant  \| K\|_F/\sqrt{\lambda\gamma}$, or uniform, leading to $\bar{L}(\pi) \leqslant \sqrt{\max\{n,d\}} \| K \|_{\max}/\sqrt{\lambda\gamma}$.
For the individual formulation where we select random elements, we consider
$\pi_{jk} =  { K_{jk}^2 }/{ \| K \|_F^2}$, leading to $\bar{L}(\pi) \leqslant    \sqrt{ \max \{n,d\} }\| K\|_F/\sqrt{\lambda\gamma}$,  or uniform, leading to
$\bar{L}(\pi) \leqslant \sqrt{nd} \| K \|_{\max}/\sqrt{\lambda\gamma}$
 (in these situations, it is important to select several elements simultaneously, which our analysis supports). 

We characterize convergence with the quantity $\varepsilon = \Omega(x-x_\ast,y-y_\ast)^2 / \Omega(x_0-x_\ast,y_0-y_\ast)^2$, where $(x_0,y_0)$ is the initialization of our algorithms (typically $(0,0)$ for bilinear saddle-points). In Table~\ref{table:summary}  we give a summary of the complexity of all  algorithms discussed in this paper: we recover the same type of speed-ups as for convex minimization. A few points are worth mentioning:

  \vspace*{-.211cm}

\BIT
\item[--] Given the bounds between the various norms on $K$ in \eq{norms}, SAGA/SVRG with non-uniform sampling always has  convergence bounds superior to SAGA/SVRG with uniform sampling, which is always superior to batch forward-backward. Note however, that in practice,  SAGA/SVRG with uniform sampling may be inferior to accelerated batch method (see \mysec{experiments}). 
\item[--] Accelerated SVRG with non-uniform sampling is the most efficient method, which is confirmed in our experiments. Note that if $n=d,$ our bound is better than or equal to accelerated forward-backward,   
	in exactly the same way than for regular convex minimization. There is thus a formal advantage for variance reduction. 

\EIT

\begin{table}

\vspace*{-.1cm}

\renewcommand{\tabcolsep}{0.13cm}
\begin{center}
\begin{tabular}{|l|rll|}
\hline
Algorithms & & Complexity &   \\
\hline
Batch FB &  $\log(1 / {\varepsilon})\  \times \big(\!$ &  $nd    +  nd \| K\|^2_{\rm op} / ( \lambda \gamma) \textcolor{white}{\Big|} $  & $\big)$\\[-.1cm]
Batch FB-accelerated    &  $\log(1 / {\varepsilon})\  \times \big(\!$ &  $nd  + nd \| K\|_{\rm op} / \sqrt{\lambda \gamma} )  \textcolor{white}{\Big|} $& $\big)$\\ 
\hline
Stochastic FB-non-uniform  &  $ (1 / {\varepsilon})\ \times \big(\!$ &  $  \max \{n,d\}   \| K\|_F^2  / ( \lambda \gamma)   \textcolor{white}{\Big|}  $ & $\big)$   \\ [-.1cm]
Stochastic FB-uniform  &  $ (1 / {\varepsilon})\ \times \big(\!$ &  $  nd \| K\|_{\rm max}^2  / ( \lambda \gamma)  \textcolor{white}{\Big|}   $   & $\big)$  \\ 
\hline
SAGA/SVRG-uniform   &  $\log(1 / {\varepsilon})\ \times \big(\!$ &  $
nd + nd \|K\|_{\rm max}^2  / ( \lambda \gamma)  \textcolor{white}{\Big|}  \!\!\!\!
   $ & $\big)$ \\[-.1cm]
SAGA/SVRG-non-uniform   &  $\log(1 / {\varepsilon})\ \times \big(\!$ &  $
nd + \max \{n,d\} \|K\|_F^2  / ( \lambda \gamma)    \textcolor{white}{\Big|} 
  $ & $\big)$\\[-.1cm]
SVRG-non-uniform-accelerated    &  $\log(1 / {\varepsilon})\ \times \big(\!$ &  $ nd + 
 \sqrt{ nd   \max \{n,d\} } \| K \|_F / \sqrt{\lambda \gamma}\textcolor{white}{\Big|} 
    $ & $\big)$\\ 
\hline
\end{tabular}
\end{center}
\caption{Summary of convergence results for the strongly $(\lambda,\gamma)$-convex-concave bilinear saddle-point problem with matrix $K$ and individual splits (and $n+d$ updates per iteration). For factored splits (little difference), see Appendix~\ref{app:factored}. For accelerated SVRG, we omitted the logarithmic term (see \mysec{acc}).
  \label{table:summary}}
\vspace*{-.5cm}
\end{table}

  \vspace*{-.211cm}

\section{SVRG: Stochastic Variance Reduction for Saddle Points}
 \label{sec:svrg}
 
 \vspace*{-.111cm}
 
Following~\cite{svrg,xiao2014proximal}, we consider a stochastic-variance reduced   estimation of   the finite sum
$ B(x,y )
= \sum_{i \in \I} B_i (x ,y  )$.
This is achieved by assuming that we have an iterate $(\tilde{x},\tilde{y})$ with a known value of $B(\tilde{x},\tilde{y})$, and we consider the estimate of $B(x,y)$:
\[ \textstyle
B(\tilde{x},\tilde{y}) + \frac{1}{\pi_i} B_i(x,y) - \frac{1}{\pi_i} B_i(\tilde{x},\tilde{y}),
\]
which has the correct expectation when $i$ is sampled from $\I$ with probability $\pi$, but with a reduced variance. Since we need to refresh $(\tilde{x}, \tilde{y})$ regularly, the algorithm works in epochs (we allow to sample $m$ elements per updates, i.e., a mini-batch of size $m$), with an algorithm that shares the same structure than SVRG for convex minimization; see Algorithm \ref{algo:svrg}. Note that we provide an explicit number of iterations per epoch, proportional to $( L^2 +   3   \bar{L}^2 / m )$.
We have the following theorem, shown in Appendix~\ref{app:svrg} (see also a  discussion of the proof  in \mysec{monotone}).
\begin{theorem}
\label{theo:svrg}
Assume (A)-(B)-(C). After $v$ epochs of  Algorithm \ref{algo:saga}, we have:
\[
\E \big[ \Omega(x_v- x_\ast, y_v - y_\ast) ^2  \big] \leqslant (3/4)^v \Omega(x_0 - x_\ast, y_0 - y_\ast) ^2 .
\]
\end{theorem}
The computational complexity to reach precision $\varepsilon$ is proportional to
$
\big[ T(B) + (m L^2 + \bar{L}^2)  \max_{i \in \I} T(B_i) + (1+ L^2 + \bar{L}^2/m) T_{\rm prox}(M) \big]   \log \frac{1}{\varepsilon}
$. Note that by taking the mini-batch size $m$ large, we can alleviate the complexity of the proximal operator ${\rm prox}_M$ if too large. Moreover, if $L^2$ is too expensive to compute, we may replace it by $\bar{L}^2$ but with a worse complexity bound.

\begin{algorithm}
\caption{SVRG: Stochastic Variance Reduction for Saddle Points} \label{algo:svrg}
\begin{algorithmic}

\REQUIRE Functions $(K_i)_i$, $M$, probabilities $(\pi_i)_i$, smoothness $\bar{L}(\pi)$ and $L$, iterate $(x,y)$\\
\hspace*{.69cm} number of epochs $v$, number of updates per iteration (mini-batch size) $m$
\STATE Set $\sigma=  \big[  L^2 +  3\bar{L}^2/m\big]^{-1}$ 
\FOR{$u$ = $1$ to $v$ }
\STATE Initialize $(\tilde{x},\tilde{y}) = (x,y)$ and compute $B(\tilde{x},\tilde{y})$
    \FOR{$k$ = $1$ to $\log 4 \times ( L^2 +   3   \bar{L}^2 / m ) $ }
    \STATE Sample $i_1,\dots,i_m \in \I$ from the probability vector $(\pi_i)_i$ with replacement
         \STATE $
(x,y) \leftarrow  {\rm prox}_{M}^\sigma \big[ (x,y) - \sigma \DD \big( B(\tilde{x},\tilde{y}) + \frac{1}{m} \sum_{k=1}^m  \big\{ \frac{1}{\pi_{i_k}} B_{i_k}(x,y) - \frac{1}{\pi_{i_k}} B_{i_k}(\tilde{x},\tilde{y}) \big\}\big)\big]
$
    \ENDFOR
\ENDFOR
\ENSURE Approximate solution $(x,y)$
\end{algorithmic}

\vspace*{-.1cm}

\end{algorithm}

\textbf{Bilinear saddle-point problems.}
When using a mini-batch size $m=1$ with the factored updates, or $m=n+d$ for the individual updates, we get the same complexities proportional to $[
nd + \max \{n,d\} \|K\|_F^2  / ( \lambda \gamma)    
]\log(1 / {\varepsilon}) $ for  non-uniform sampling, which improve significantly over (non-accelerated) batch methods (see Table~\ref{table:summary}).

\section{SAGA: Online Stochastic Variance Reduction for Saddle Points}
 
 \vspace*{-.111cm}
 
\label{sec:saga}
Following~\cite{defazio2014saga}, we consider a stochastic-variance reduced   estimation of  
$B(x,y )
= \sum_{i \in \I} B_i (x ,y  ). $
This is achieved by assuming that we store values $g^i = B_i (x^{{\rm old}(i)},y^{{\rm old}(i)})$ for an old iterate 
$(x^{{\rm old}(i)},y^{{\rm old}(i)})$, and we consider the estimate of $B(x,y)$:
\[ \textstyle
\sum_{j \in \I} g^j + \frac{1}{\pi_i} B_i(x,y) - \frac{1}{\pi_i} g^i,
\]
which has the correct expectation when $i$ is sampled from $\I$ with probability $\pi$. At every iteration, we also refresh the operator values $g^i \in \rb^{n + d}$, for the same index $i$ or with a new  index $i$ sampled uniformly at random. This leads to Algorithm \ref{algo:saga}, and we have the following theorem showing linear convergence, proved in Appendix~\ref{app:saga}. Note that for bi-linear saddle-points, the initialization at $(0,0)$ has zero cost (which is not possible for convex minimization).
\begin{theorem}
\label{theo:saga}
Assume (A)-(B)-(C). After $t$ iterations of  Algorithm \ref{algo:saga} (with the option of resampling when using non-uniform sampling), we have:
\[ \textstyle
\E \big[ \Omega(x_t - x_\ast, y_t - y_\ast) ^2  \big] \leqslant  2 
\big( 1 -     (  \max \{ \frac{3|\I|}{ 2m}, 1 +  \frac{L^2}{\mu^2} + 3 \frac{\bar{L}^2}{ m  \mu^2} \} )^{-1}  \big)^t \ \Omega(x_0 - x_\ast, y_0 - y_\ast) ^2.\]
\end{theorem}
\textbf{Resampling or re-using the same gradients.}  For the bound above to be valid for non-uniform sampling, like for convex minimization~\cite{resa}, we need to resample $m$ operators after we make the iterate update. In our experiments, following~\cite{resa}, we considered a mixture of uniform and non-uniform sampling, without the resampling step.

\textbf{SAGA vs. SVRG.} The difference between the two algorithms is the same as for convex minimization (see, e.g.,~\cite{harikandeh2015stopwasting} and references therein), that is SVRG   has no storage, but works in epochs and requires slightly more accesses to the oracles, while SAGA is a pure online method with fewer parameters but requires some storage (for bi-linear saddle-point problems, we only need to store $O(n\!+\!d)$ elements for the factored splits, while we need $O(dn)$ for the individual splits). Overall they have the same running-time complexity for individual splits; for factored splits, see Appendix~\ref{app:factored}.

\textbf{Factored splits.} When using factored splits, we need to store the two parts of the operator values separately and update them independently, leading in Theorem~\ref{theo:saga} to replacing
 $|\I|$ by $\max\{|\mathcal{J}|, |\mathcal{K}|\}$.

\begin{algorithm}
\caption{SAGA: Online Stochastic Variance Reduction for Saddle Points} \label{algo:saga}
\begin{algorithmic}

\REQUIRE Functions $(K_i)_i$, $M$, probabilities $(\pi_i)_i$, smoothness $\bar{L}(\pi)$ and $L$, iterate $(x,y)$ \\
\hspace*{.69cm} number of iterations $t$, number of updates per iteration (mini-batch size) $m$
\STATE Set $\sigma=  
\big[ \max \{ \frac{3|\I|}{2 m} -1 , L^2 + 3 \frac{\bar{L}^2}{ m   }  \} \big]^{-1}$
\STATE Initialize $g^i = B_i (x,y)$ for all $i \in \I$ and $G= \sum_{i \in \I} g^i$
\FOR{$u$ = $1$ to $t$ }
      \STATE Sample $i_1,\dots,i_m \in \I$ from the probability vector $(\pi_i)_i$ with replacement
 \STATE  Compute $h_{k} = B_{i_k}(x,y) $ for $k \in \{1,\dots,m\}$
        \STATE $
(x,y) \leftarrow  {\rm prox}_{M}^\sigma \big[ (x,y) - \sigma \DD \big( G + \frac{1}{m}\sum_{k=1}^m \big\{
\frac{1}{\pi_{i_k}} h_k  - \frac{1}{\pi_{i_k}}g^{i_k} \big\}\big)\big]
$
      \STATE (optional) Sample $i_1,\dots,i_m \in \I$ uniformly with replacement
 \STATE   (optional) Compute $h_{k} = B_{i_k}(x,y) $ for $k \in \{1,\dots,m\}$
\STATE Replace $G \leftarrow G - \sum_{k=1}^m \{ g^{i_k} - h_k \}$ and $g^{i_k} \leftarrow h_k$ for $k \in \{1,\dots,m\}$
\ENDFOR
\ENSURE Approximate solution $(x,y)$
\end{algorithmic}

\vspace*{-.1cm}

\end{algorithm}

\section{Acceleration}
 
 \vspace*{-.111cm}
 
\label{sec:acc}
Following the ``catalyst'' framework of~\cite{lin2015universal}, we  consider a sequence of saddle-point problems with added regularization; namely, given $(\bar{x},\bar{y})$, we use SVRG to solve approximately
\BEQ
\label{eq:ppa}
\min_{x \in \rb^d} \max_{y \in \rb^n} 
\textstyle
K(x,y) + M(x,y) + \frac{\lambda \tau}{2} \| x - \bar{x}\|^2 -
\frac{\gamma\tau }{2} \| y - \bar{y}\|^2,
\EEQ
for well-chosen $\tau$ and $(\bar{x},\bar{y})$.
The main iteration of the algorithm differs from the original SVRG by the presence of the iterate $(\bar{x},\bar{y}) $, which is updated regularly (after a precise number of epochs), and different step-sizes (see details in Appendix~\ref{app:accsvrg}).
The complexity to get an approximate solution of \eq{ppa} (forgetting the complexity of the proximal operator and for a single update), up to logarithmic terms, is proportional, to $T(B) + \bar{L}^2 ( 1 + \tau)^{-2} \max_{i \in \I} T(B_i)$. 

The key difference with the convex optimization set-up is that the analysis is simpler, without the need for Nesterov acceleration machinery~\cite{nest2004} to define a good value of $(\bar{x},\bar{y})$; indeed, the solution of \eq{ppa} is one iteration of the proximal-point algorithm, which is known to converge linearly~\cite{rockafellar1976monotone} with rate $ (1  +  \tau^{-1})^{-1 }
= ( 1 - \frac{1}{1+\tau})$. Thus the overall complexity is up to logarithmic terms equal to
$T(B) ( 1+ \tau) + \bar{L}^2 ( 1 + \tau)^{-1} \max_{i \in \I} T(B_i)$. The trade-off in $\tau$ is optimal for
$1+ \tau = \bar{L} \sqrt{   {\max_{i \in \I} T(B_i)}/ {T(B)}}$, showing that there is a potential acceleration when 
$\bar{L} \sqrt{   {\max_{i \in \I} T(B_i)}/{T(B)}} \geqslant 1$, leading to a complexity
$\bar{L}   \sqrt{ T(B)   \max_{i \in \I} T(B_i ) }$.

  Since  the SVRG algorithm already works in epochs, this leads to a simple modification where every $ \log( 1 + \tau)$ epochs, we change the values of $(\bar{x},\bar{y})$. See Algorithm~\ref{algo:svrg-acc} in Appendix~\ref{app:accsvrg}. Moreover, we can adaptively update $(\bar{x},\bar{y})$ more aggressively to speed-up the algorithm.

The following theorem gives the convergence rate of the method (see proof in Appendix~\ref{app:accsvrg}). With the value of $\tau$ defined above (corresponding to $\tau = \max \big\{ 0, \frac{\|K\|_F}{\sqrt{\lambda \gamma}} \sqrt{\max\{n^{-1},d^{-1}\}}-1 \big\}$ for bilinear problems), we get the complexity
$\bar{L}   \sqrt{ T(B)   \max_{i \in \I} T(B_i ) }$, up to the logarithmic term $\log(1+\tau)$. For bilinear problems, this provides a significant  acceleration, as shown in Table~\ref{table:summary}.
\begin{theorem}
\label{theo:acc}
Assume (A)-(B)-(C). After $v$ epochs of  Algorithm \ref{algo:svrg-acc}, we have, for any positive $v$:
\[ \textstyle
\E \big[ \Omega(x_v - x_\ast, y_v - y_\ast) ^2  \big] \leqslant  \big( 1 -   \frac{1}{\tau + 1}\big)^v \ \Omega(x_0 - x_\ast, y_0 - y_\ast) ^2.\]
\end{theorem}
While we provide a proof only for SVRG, the same scheme should work for SAGA. Moreover, the same idea also applies to the batch setting (by simply considering $|\I|=1$, i.e., a single function), leading to an acceleration, but now valid for all functions $K$ (not only bilinear).

\section{Extension to Monotone Operators}
 
 \vspace*{-.111cm}
 
\label{sec:monotone}
In this paper, we have chosen to focus on saddle-point problems because of their ubiquity in machine learning. However, it turns out that our algorithm  and, more importantly, our analysis extend to all set-valued \emph{monotone operators}~\cite{bauschke2011convex,ryu}.
 We  thus consider a maximal strongly-monotone operator $A$ on a Euclidean space $\mathcal{E}$, as well as a finite family of Lipschitz-continuous (not necessarily monotone) operators $B_i$, $i \in \I$, with $B = \sum_{i \in \I} B_i$ monotone. Our algorithm then finds the zeros of $ A + \sum_{i \in \I} B_i = A + B$, from the knowledge of the resolvent (``backward'') operator $( \idm + \sigma A )^{-1}$ (for a well chosen $\sigma>0$) and the forward operators $B_i$, $i \in \I$. Note the difference with~\cite{raguet2013generalized}, which requires each $B_i$ to be monotone with a known resolvent and $A$ to be monotone and single-valued. There several interesting examples (on which our algorithms apply):
  
  \vspace*{-.211cm}

 \BIT
 \item[--] \textbf{Saddle-point problems}: 
 We assume for simplicity that $\lambda = \gamma = \mu$ (this can be achieved by a simple change of variable).
If we denote $B(x,y) = ( \partial_x K(x,y),  -\partial_y K(x,y) )$ and the multi-valued operator $A(x,y) = ( \partial_x M(x,y),  -\partial_y M(x,y) )$, then the proximal operator ${\rm prox}_M^\sigma$ may be written as $(\mu \idm +\sigma A)^{-1}(\mu x,\mu y)$, and we recover exactly our framework from \mysec{comp}.

  \vspace*{-.07cm}

 \item[--] \textbf{Convex minimization}: $A = \partial g$ and $B_i = \partial f_i$ for a strongly-convex function $g$ and smooth functions $f_i$: we recover proximal-SVRG~\cite{xiao2014proximal} and SAGA~\cite{defazio2014saga}, to minimize $\min_{z \in \mathcal{E}} g(z) + \sum_{i \in \I} f_i(z)$. However, this is a situation where the operators $B_i$ have an extra property called co-coercivity~\cite{zhu1996co}, which we are not using because it is not satisfied for saddle-point problems. The extension of SAGA and SVRG to monotone operators was proposed earlier by~\cite{davis2016smart}, but only co-coercive operators are considered, and thus only convex minimization is considered (with important extensions beyond plain SAGA and SVRG), while our analysis covers a much broader set of problems. In particular, the step-sizes obtained with co-coercivity lead to divergence in the general setting.

Because we do not use  co-coercivity, applying our results directly to convex minimization, we would get slower rates, while, as shown in \mysec{examples},  they can be easily cast as a saddle-point problem  if the proximal operators of the functions $f_i$ are known, and we then get the same rates than existing fast techniques which are dedicated to this problem~\cite{roux2012stochastic,svrg,defazio2014saga}.

   \vspace*{-.07cm}

\item[--] \textbf{Variational inequality problems}, which are notably common in game theory (see, e.g.,~\cite{harker1990finite}). 
 \EIT

\section{Experiments}
\label{sec:experiments}
 
 \vspace*{-.111cm}
 
 We consider binary classification   problems with design matrix $K$ and label vector  in $ \{-1,1\}^n$, a non-separable strongly-convex regularizer with an efficient proximal operator (the sum of the squared norm $\lambda \| x\|^2 / 2$ and  the clustering-inducing term  $\sum_{i  \neq j} |x_i - x_j|$, for which the proximal operator may be computed in $O(n \log n)$ by isotonic regression~\cite{oscar}) and a non-separable smooth loss (a surrogate to the area under the ROC curve, defined as proportional to $\sum_{i_+ \in I_+} \sum_{i_- \in I_-} ( 1 - y_i + y_j)^2$, where $I_+$/$I_-$ are sets with positive/negative labels, for a vector of prediction $y$, for which an efficient proximal operator may be computed as well, see Appendix~\ref{app:auc}).

Our upper-bounds depend  on the ratio $\| K\|_F^2 / ( \lambda \gamma)$ where $\lambda$ is the regularization strength and $\gamma \approx n$ in our setting where we minimize an average risk. Setting $\lambda = \lambda_0 = \| K \|_F^2 / n^2$ corresponds to a regularization proportional to the average squared radius of the data divided by $1/n$ which is standard in this setting~\cite{roux2012stochastic}. We also experiment with  smaller regularization (i.e., $\lambda/\lambda_0 = 10^{-1}$), to make the problem more ill-conditioned (it turns out that the corresponding testing losses are sometimes slightly better). We consider two datasets, \texttt{sido} ($n=10142$, $d=4932$, non-separable losses and regularizers presented above) and \texttt{rcv1} ($n=20242 $, $d=47236$, separable losses and regularizer described in Appendix~\ref{app:results}, so that we can compare with SAGA run in the primal). We report below the squared distance to optimizers which appears in our bounds, as a function of the number of passes on the data (for more details and experiments with primal-dual gaps and testing losses, see Appendix~\ref{app:results}).  Unless otherwise specified, we always use non-uniform sampling.  

\begin{figure}[H]

\vspace*{-.3cm}

\hspace*{-.00cm}\includegraphics[scale=.38]{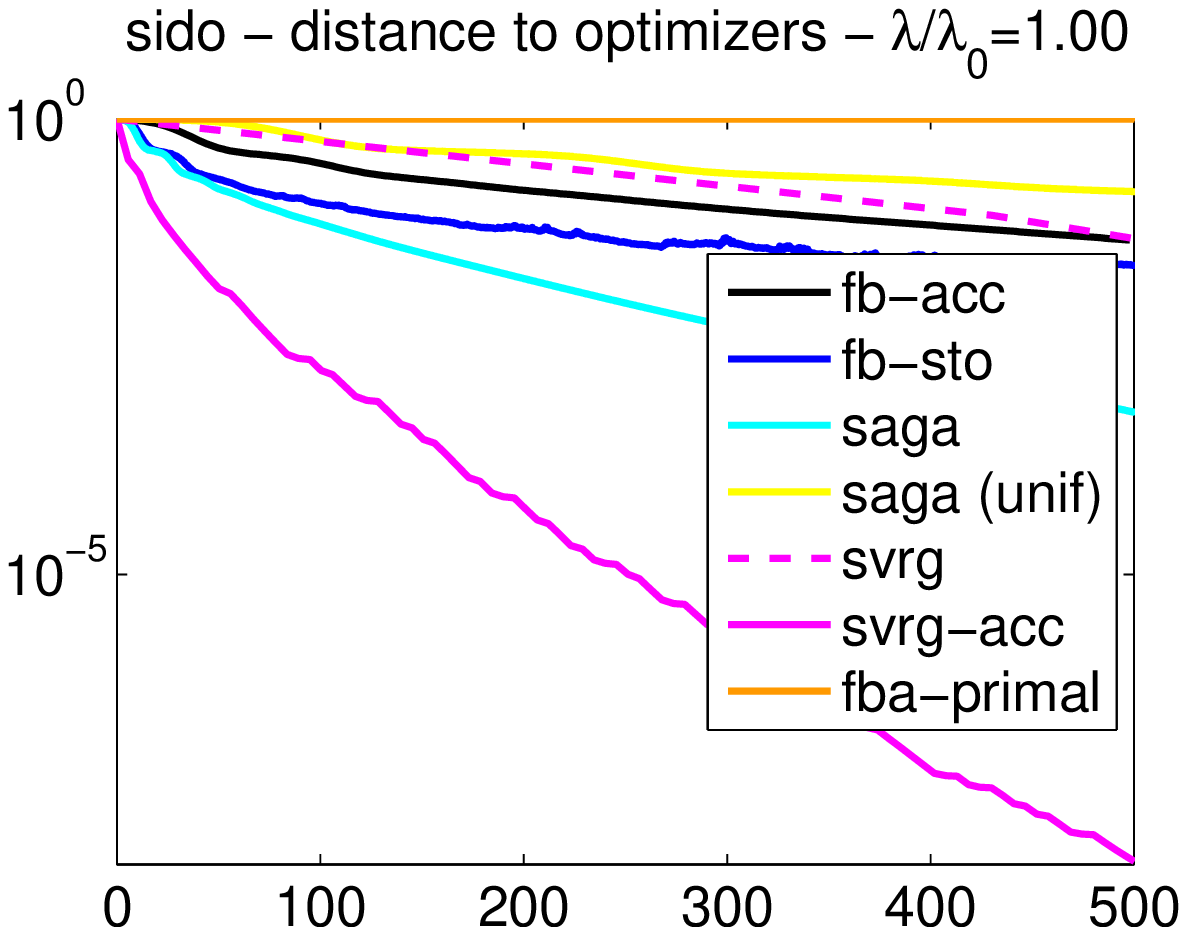}
\hspace*{-0.1cm}
\includegraphics[scale=.38]{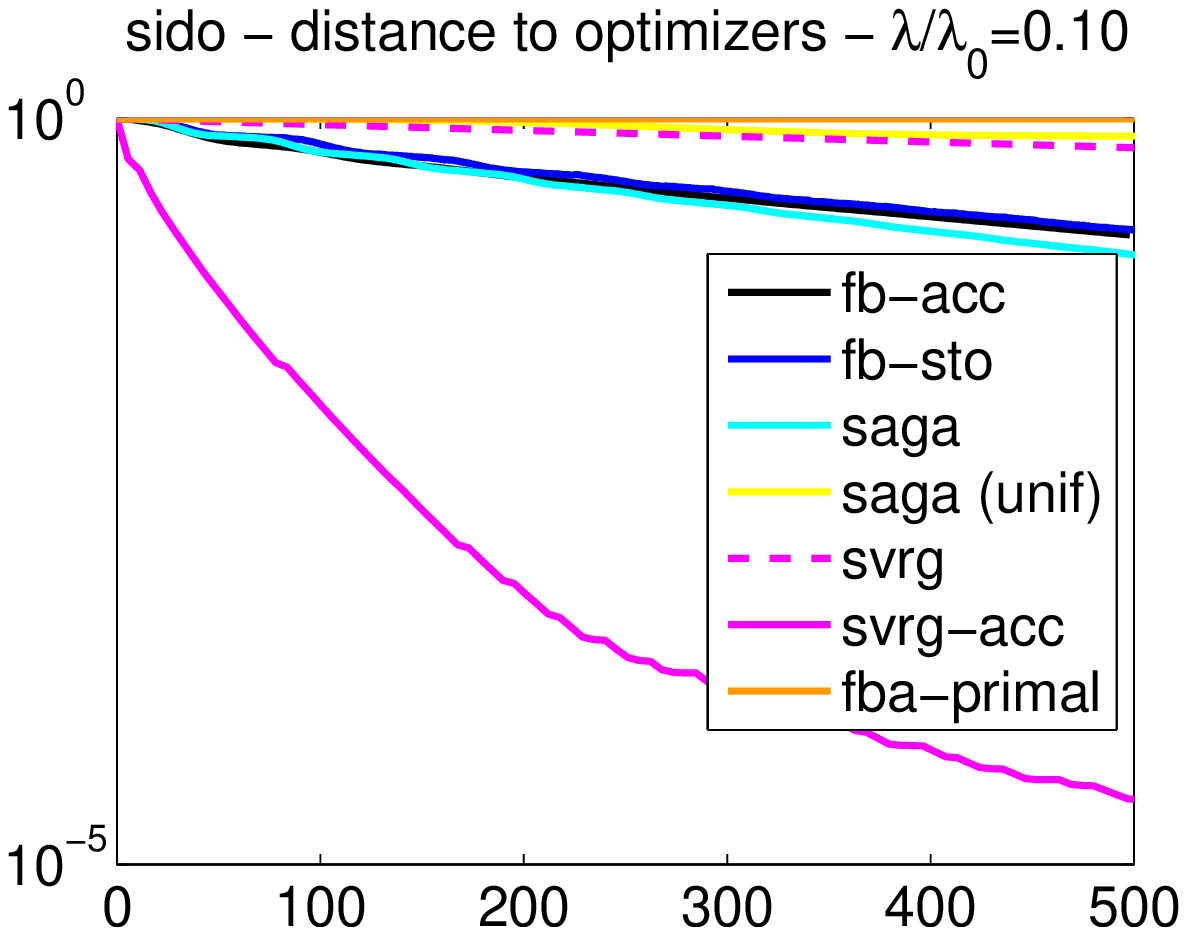}
\hspace*{-0.1cm}
\includegraphics[scale=.38]{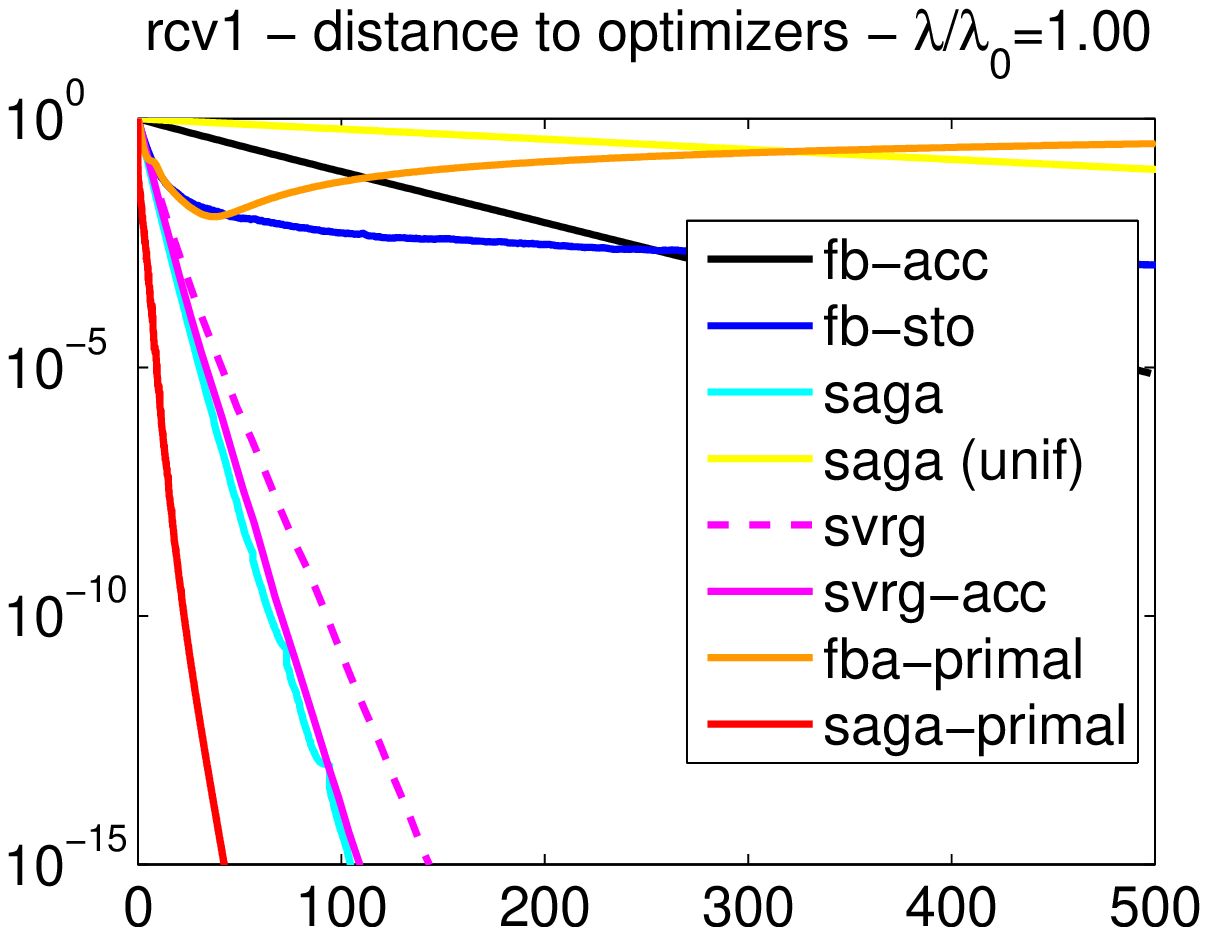}
\hspace*{-3.5cm}

\vspace*{-.4cm}

\end{figure}

We   see that uniform sampling for SAGA does not improve on batch methods, SAGA and accelerated SVRG (with non-uniform sampling) improve significantly over the existing methods, with a stronger gain for the accelerated version for ill-conditioned problems (middle vs.~left plot). On the right plot, we compare to primal methods on a separable loss, showing that primal methods (here ``fba-primal'', which is Nesterov acceleration) that do not use separability (and can thus be applied in all cases) are inferior, while SAGA run on the primal remains  faster (but cannot be applied for non-separable losses).

 \vspace*{-.111cm}

\section{Conclusion}

 \vspace*{-.211cm}

We proposed the first linearly convergent incremental gradient algorithms for saddle-point problems, which improve both in theory and practice over existing batch or stochastic algorithms. While we currently need to know the strong convexity-concavity constants, we plan to explore in future work adaptivity to these constants like already obtained for convex minimization~\cite{defazio2014saga}, paving the way to an analysis without strong  convexity-concavity.

\newpage
{\small
\bibliography{sagasaddle}
}
\newpage

\appendix

\begin{center}

{\Large Stochastic Variance Reduction Methods
 for Saddle-Point Problems \\[.2cm]
P.~Balamurugan and F. Bach}
 
{\Large \emph{Supplementary material - NIPS 2016} }

\end{center}

\vspace*{1cm}

\section{Formalization through Monotone Operators}

 \vspace*{-.111cm}

Throughout the proofs, we will consider only maximal monotone operators on a Euclidean space $\mathcal{E}$, that is $A$ is assumed to be a $\mu$-strongly monotone (corresponding to $M$ for saddle-points) and potentially set-valued, while $B$ is monotone and $L$-Lipschitz-continuous with respect to the Euclidean norm (and hence single-valued). For an introduction to monotone operators, see~\cite{bauschke2011convex,ryu}.

For simplicity, in this appendix, we will only consider a single-valued operator $A$ (noting that the proof extends to any set-valued operator $A$), and we will mostly focus here on the monotonicity properties (noting that the ``maximal'' property can be treated rigorously~\cite{bauschke2011convex}, in particular to ensure that the resolvent operator is defined everywhere). An operator is monotone if and only if for all $(z,z')$, $ ( A(z) - A(z') )^\top ( z - z') \geqslant 0 $. The most basic example is the subdifferential of a convex function. In this paper, we focus on saddle-point problems.

\textbf{Application to saddle-point problems.}
For the saddle-point problems defined in \mysec{comp} of the main paper, where we have $z=(x,y)$, we need to make a change of variable because of the two potentially different scaling factors $\lambda$ and $\gamma$. We consider the operators
\BEAS
B(x,y) & = &  ( \lambda^{-1/2} \partial_x K( \lambda^{-1/2} x, \gamma^{-1/2} y), -\gamma^{-1/2} \partial_y K( \lambda^{-1/2} x, \gamma^{-1/2} y)) \\
A(x,y) & = &  ( \lambda^{-1/2} \partial_x M( \lambda^{-1/2} x, \gamma^{-1/2} y), -\gamma^{-1/2} \partial_y M( \lambda^{-1/2} x, \gamma^{-1/2} y)).
\EEAS
The solutions of $A(x,y) + B(x,y) = 0$ are exactly the solutions of the problem in \eq{saddle}, rescaled by $\lambda^{1/2}$ and $\gamma^{1/2}$. Moreover, the operator $A$ is  $\mu$-monotone with $\mu=1$, i.e., for any $z,z'$, we have $ ( A(z) - A(z') )^\top ( z - z') \geqslant \|z-z'\|^2$.
Finally, our definition of the smoothness constants for $B$ and $B_i$ in the main paper, exactly leads to a Lipschitz-constant of $L$ with respect to the natural Euclidean norm (a similar result holds for the constant $\bar{L}(\pi)$ defined later). Moreover,  convergence results in the Euclidean norm here transfer to convergence results in the norm $\Omega$ defined in the main paper. Note that because of our proofs through operators, it is not easily possible to get bounds on the primal and dual gaps.

\textbf{Properties of monotone operators and resolvents.}
Given a maximal monotone operator $A$, we may define its \emph{resolvent} operator as $z' = (\idm + \sigma A)^{-1}(z)$, which is defined as finding the unique $z'$ such that $ z' + \sigma A(z') = z$. When $A$ is the operator associated to the saddle-point function $M$ as described above, then the resolvent operator is exactly the proximal operator of $M$ defined in \eq{prox} of the main paper. Note that care has to be taken with the scaling factors $\lambda$ and $\gamma$.

We will use the following properties (on top of Lipschitz-continuity) \cite{bauschke2011convex,ryu}:
\BIT
\item[--] Monotonicity property:  for any $(z,z')$, $( B(z) - B(z'), z-z') \geqslant 0$.
\item[--] Contractivity of the resolvent operator for $A$ $\mu$-strongly-monotone: for any $(z,z')$, $\| (\idm + \sigma A)^{-1}(z) - (\idm + \sigma A)^{-1}(z') \| 
\leqslant (1 + \sigma \mu)^{-1} \| z - z'\|$.
\item[--] Firm non-expansiveness of the resolvent: for any $(z,z')$, $\| (\idm + \sigma A)^{-1}(z) - (\idm + \sigma A)^{-1}(z') \|^2 
\leqslant (1 + \sigma \mu)^{-1} ( z - z')^\top \big( (\idm + \sigma A)^{-1}(z) - (\idm + \sigma A)^{-1}(z') \big)$.
\EIT

Moreover, given our strong-monotonicity assumption, $A+B$ has a unique zero $z_\ast \in \mathcal{E}$.

Finally in order to characterize the running-times, we will consider the complexity $T_{\rm fw}(B)$ of computing the operator $B$ and the complexity $T_{\rm bw}(A)$ to compute the resolvent of $A$. For saddle-point problems, these correspond to $T(B)$ and $T_{\rm prox}(M)$ from the main paper.

\section{Proof for Deterministic Algorithms}

 \vspace*{-.111cm}

All proofs in this section will follow the same principle, by showing that at every step of our algorithms, a certain function (a ``Lyapunov'' function) is contracted by a factor strictly less than one. For the forward-backward algorithm, this will be the distance to optimum $\|z_t - z_\ast\|^2$; while for the accelerated version, it will be different.


\subsection{Forward-backward algorithm}

\vspace*{-.051cm}

\label{app:fb}
We consider the iteration $z_t = ( I + \sigma A)^{-1} (z_{t-1} - \sigma B (z_{t-1}))$, with $B$ being monotone $L$-Lipschitz-continuous and $A$ being $\mu$-strongly monotone. The optimum $z_\ast$ (i.e., the zero of $A+B$) is invariant by this iteration.
Note that this is the analysis of~\cite{chen1997convergence} and that we could improve it by putting some of the strong-monotonicity in the operator $B$ rather than in $A$.

We have:
\BEAS
 & & \| z_t - z_\ast \|^2\\
  & \leqslant & \frac{1}{(1+\sigma \mu)^2} \| z_{t-1} -z_\ast - \sigma ( B (z_{t-1}) - B (z_\ast)) \|^2 
\mbox{ by contractivity of the resolvent,}\\
& = &  \frac{1}{(1+\sigma \mu)^2}  \Big[
\| z_{t-1} -z_\ast\|^2 - 2
\sigma (z_{t-1} -z_\ast)^\top  ( B (z_{t-1}) - B (z_\ast))  + \sigma^2 \| B (z_{t-1}) - B (z_\ast)\|^2
\Big] \\
& \leqslant & \frac{1}{(1+\sigma \mu )^2}   ( 1 + \sigma^2 L^2 ) \| z_{t-1} -z_\ast\|^2
\mbox{ by monotonicity of   and Lipschitz-continuity of } B,
\\
& \leqslant & \Big( \frac{   1 + \sigma^2 L^2 }{(1+\sigma \mu)^2} \Big)^t \| z_0 - z_\ast\|^2, \mbox{ by applying the recursion } t \mbox{ times}.
\EEAS
Thus we get linear (i.e., geometric) convergence as soon as $1+ \sigma^2 L^2 <  ( 1 + \sigma \mu)^2$. If we consider
$\eta = \frac{\sigma \mu }{1+\sigma \mu} \in [0,1)$, and the rate above becomes equal to:
\[
 \frac{   1 + \sigma^2 L^2 }{(1+\sigma \mu)^2} = ( 1 -\eta)^2 + \eta^2 \frac{L^2}{\mu^2} = 1 - 2 \eta + \eta^2 ( 1 + \frac{L^2}{\mu^2}),
\]
thus the algorithm converges if $\eta < \frac{2}{1 + \frac{L^2}{\mu^2}}$, and with $\eta = \frac{1}{1 + \frac{L^2}{\mu^2}}$ which corresponds to $\sigma = \frac{1}{\mu} \frac{ \eta}{1-\eta} = \frac{\mu}{L^2}$, we get a linear convergence rate with constant $ 1 - \eta = \frac{L^2}{\mu^2 + L^2}$. 

Thus the complexity to reach the precision $\varepsilon \times \| z_0 - z_\ast\|^2$ in squared distance to optimum $\| z_t - z_\ast\|^2$ is equal to
$ \big( 1 + \frac{L^2}{\mu^2} )\big[ T_{\rm fw}(B) + T_{\rm bw}(A) \big] \log \frac{1}{\varepsilon}$.

Note that we obtain a slow convergence when applied to convex minimization, because we are not using any co-coercivity of $B$, which would lead to a rate $(1-\mu/L)$~\cite{zhu1996co}. Indeed, co-coercivity means that $\| B(z) - B(z') \|^2 \leqslant
L ( B(z)-B(z'))^\top (z-z')$, and this allows to replace above the term $1 + \sigma^2 L^2$ by $1 $ if $\sigma \leqslant 2/L$, leading to linear convergence rate with constant $( 1 + \mu/L)^{-2} \approx 1 - 2 \mu / L$.

\subsection{Accelerated forward-backward algorithm}

\vspace*{-.051cm}

\label{app:acc}
We consider the iteration $z_t = ( I + \sigma A)^{-1} (z_{t-1} - \sigma B [ z_{t-1} + \theta( z_{t-1} - z_{t-2} ) ])$, with $B$ being monotone $L$-Lipschitz-continuous and \emph{linear}, and $A$ being $\mu$-strongly monotone. 
Note that this is an extension of the analysis of~\cite{chambolle2011first} to take into account the general monotone operator situation. Again $z_\ast$ is a fixed-point of the iteration.

Using the firm non-expansiveness of the resolvent operator, we get, with $\eta = \frac{\sigma \mu}{1 + \sigma \mu}$, and then using the linearity of $B$:
 \BEAS
 \| z_ t -z_\ast\|^2 & \leqslant & \frac{1}{1 + \sigma \mu }(z_t - z_\ast)^\top \Big[
 z_{t-1} - z_\ast  - \sigma B [ z_{t-1} - z_\ast + \theta ( z_{t-1} - z_{t-2})  ] \Big] \\
 & = & 
 (z_t - z_\ast)^\top \Big[
 (1-\eta) ( z_{t-1} - z_\ast)   - \frac{\eta}{\mu} B [ z_{t-1} - z_\ast + \theta ( z_{t-1} - z_{t-2})  ] \Big] \\
& = & - \frac{1 - \eta}{2} \| z_t - z_{t-1} \|^2 + \frac{1-\eta}{2} \| z_{t} -z_\ast\|^2
+   \frac{1-\eta}{2} \| z_{t-1} -z_\ast \|^2  \\
& & \hspace*{4cm}
- \frac{\eta}{\mu}  (z_t - z_\ast)^\top  B [ z_{t-1} - z_\ast + \theta ( z_{t-1} - z_{t-2})  ]
\\
& = & - \frac{1 - \eta}{2} \| z_t - z_{t-1} \|^2 + \frac{1-\eta}{2} \| z_{t} - z_\ast \|^2
+   \frac{1-\eta}{2} \| z_{t-1} -z_\ast \|^2 \\
& & \hspace*{3cm}
-\frac{\eta}{\mu} (z_t - z_\ast)^\top B ( z_{t-1}  - z_\ast ) 
- \theta \frac{\eta}{\mu}  (z_t - z_\ast)^\top B ( z_{t-1}- z_{t-2} ) ,
\EEAS
by regrouping terms. By using the Lipschitz-continuity of $B$, we get:
\BEAS
 & & \| z_ t -z_\ast\|^2  \\
 & \leqslant & - \frac{1 - \eta}{2} \| z_t - z_{t-1} \|^2 + \frac{1-\eta}{2} \| z_{t} - z_\ast\|^2
+   \frac{1-\eta}{2} \| z_{t-1} -z_\ast\|^2
- \frac{\eta}{\mu}( z_t -z_\ast)^\top B  ( z_{t-1}  - z_t ) 
\\
& & - \theta \frac{\eta}{\mu}
(z_{t-1}  -z_\ast) ^\top B  ( z_{t-2}  - z_{t-1} ) 
+ 
 \theta \frac{\eta}{\mu}  L  \|z_t -z_{t-1} \| \| z_{t-1}  - z_{t-2}\|
\\
& \leqslant & - \frac{1 - \eta}{2} \| z_t - z_{t-1} \|^2 + \frac{1-\eta}{2} \| z_{t} - z_\ast\|^2
+   \frac{1-\eta}{2} \| z_{t-1} - z_\ast \|^2
- \frac{\eta}{\mu}( z_t -z_\ast)^\top B  ( z_{t-1}  - z_t ) \\
& & - \theta \frac{\eta}{\mu}
(z_{t-1}  -z_\ast) ^\top B  ( z_{t-2}  - z_{t-1} ) 
+ 
 \frac{\theta   L }{2} \frac{\eta}{\mu} \big[ \alpha^{-1} \|z_t -z_{t-1} \|^2  +  \alpha\| z_{t-1}  - z_{t-2}\|^2  \big],
 \EEAS
 with a constant $\alpha >0$ to be determined later.
This leads to, with $\theta = \frac{1-\eta}{1+\eta}$, and by regrouping terms:
\BEAS
& & \frac{1+\eta}{2} \| z_t -z_\ast\|^2
+ \Big(
\frac{1-\eta}{2} -  \frac{\theta \eta L }{2 \mu}  \alpha^{-1}
\Big) \|z_t -z_{t-1} \|^2 - \eta (z_t -z_\ast)^\top B  ( z_{t-1}  - z_t )  \\
& \leqslant & \frac{1-\eta}{2} \| z_{t-1} -z_\ast \|^2
+
\Big(  \frac{\alpha \eta \theta L}{2 \mu}  \Big)
 \|z_{t-1} -z_{t-2} \|^2 - \theta \frac{\eta}{\mu}  (z_{t-1}-z_\ast)^\top B ( z_{t-2}  - z_{t-1} ) 
\\
& \leqslant & \theta \bigg[ \frac{1+\eta}{2} \| z_{t-1} - z_\ast\|^2
+
\Big(  \frac{\eta \alpha  L }{2 \mu}  \Big)
 \|z_{t-1} -z_{t-2} \|^2 -   \frac{\eta}{\mu} (z_{t-1}-z_\ast)^\top B ( z_{t-2}  - z_{t-1} ) \bigg].
\EEAS
We get a Lyapunov function $\mathcal{L}: (z,z') \mapsto
\frac{1+\eta}{2} \| z -z_\ast\|^2
+ \Big(
\frac{1-\eta}{2} -  \frac{\theta \eta L }{2 \mu}  \alpha^{-1}
\Big) \|z -z' \|^2 - \eta (z -z_\ast)^\top B  ( z'  - z )  
$, such  that  $\mathcal{L}(z_t,z_{t-1})$ converges to zero geometrically, if
$
 \frac{\alpha  \eta  L}{\mu} \leqslant  {1-\eta}  -   \frac{\eta \theta L}{\mu}\alpha^{-1}
$
and
$
\left( \begin{array}{cc}
1+ \eta & -  \eta L /\mu \\
\eta L / \mu & 1 - \eta -  \eta \theta L \mu^{-1} \alpha^{-1}
\end{array} \right) \succcurlyeq 0
$.
By setting $\eta = \frac{1}{1 + 2 L / \mu}$, and thus
$\theta = \frac{1-\eta}{1+\eta} = \frac{1}{1+\mu/L}$, $\sigma = \frac{1}{\mu}\frac{\eta}{1-\eta}  =  \frac{1}{2L}$,
and $\alpha = 1$, we get the desired first property and the fact that the matrix above is greater than $\left( \begin{array}{cc}
1/2  & 0   \\
0 &  0 \\
\end{array} \right)$, which allows us to get a linear rate of convergence for $\|z_t - z_\ast\|^2 \leqslant 2 \mathcal{L}(z_t,z_{t-1})$.

\section{Proof for Existing Stochastic Algorithms}

 \vspace*{-.111cm}
 
\label{app:sto}
We follow~\cite{rosasco2014stochastic}, but with a specific step-size that leads to a simple result, which also applies to non-uniform sampling from a finite pool.
We consider the iteration $z_t = ( I + \sigma_t A)^{-1} (z_{t-1} - \sigma_t ( B z_{t-1} + C_t z_{t-1} ) )$, with $B$ being monotone $L$-Lipschitz-continuous and $A$ being $\mu$-strongly monotone, and $C_t $ a random operator (\emph{not necessarily monotone}) such that $\E C_t (z)= 0$ for all $z$. We assume that all random operators $C_t$ are independent, and we denote by~$\mathcal{F}_t$ the $\sigma$-field generated by $C_1,\dots,C_t$, i.e., the information up to time $t$.

We have with ${\rm Lip}(C_t)$ the Lipschitz-constant of $C_t$:
\BEAS
\| z_t - z_\ast \|^2 & \leqslant & \frac{1}{(1+\sigma_t \mu)^2} \| z_{t-1} -z_\ast - \sigma_t ( B (z_{t-1}) - B (z_\ast))  - \sigma_t C_t (z_{t-1}) \|^2  \\
& & \hspace*{4cm}
\mbox{ by contractivity of the resolvent,}\\
& = &  \frac{1}{(1+\sigma_t \mu)^2}  \Big[
\| z_{t-1} -z_\ast\|^2 - 2
\sigma_t (z_{t-1} -z_\ast)^\top  ( B (z_{t-1}) - B (z_\ast))  \\
& & \hspace*{0.1cm} + \sigma_t^2 \| B (z_{t-1}) - B ( z_\ast) + C_t (z_{t-1})\|^2
+ 2 \sigma_t (C_t (z_{t-1}))  ^\top ( z_{t-1} -z_\ast  )
\Big] .
\EEAS
By taking conditional expectations, we get:
\BEAS
\!\!\!\!\!\!
\E \big(
\| z_t - z_\ast \|^2 \big| \mathcal{F}_{t-1} \big)
\!\!& \leqslant & \frac{1}{(1+\sigma_t \mu )^2} \big[
  ( 1 +   \sigma_t^2 L^2 ) \| z_{t-1} -z_\ast\|^2
  +  \sigma_t^2 \E ( \| C_t (z_{t-1})\|^2 | \mathcal{F}_{t-1})
  \big] \\
  & & 
\mbox{ by monotonicity and Lipschitz-continuity of } B,
\\
& \leqslant & \frac{1}{(1+\sigma_t \mu )^2} \big[
  ( 1 +   \sigma_t^2 L^2 ) \| z_{t-1} -z_\ast\|^2
  +  2 \sigma_t^2 \E ( \| C_t (z_{\ast})\|^2 | \mathcal{F}_{t-1}) \\
  & &   \hspace*{2cm}
  + 2 \sigma_t^2 \| z_{t-1} - z_\ast\|^2 \E ( \sup_{\|z-z'\|=1} \| C_t (z) - C_t(z')\|^2 | \mathcal{F}_{t-1})
  \big]
\\
& = & \frac{1}{(1+\sigma_t \mu )^2} \big[
  ( 1 +   \sigma_t^2 L^2 ) \| z_{t-1} -z_\ast\|^2
  +  2 \sigma_t^2 \E ( \| C_t (z_{\ast})\|^2 | \mathcal{F}_{t-1}) \\
  & &   \hspace*{4cm}
  + 2 \sigma_t^2 \| z_{t-1} - z_\ast\|^2 \E ( {\rm Lip}(C_t )^2 | \mathcal{F}_{t-1})
  \big]
\\
& = & \frac{1}{(1+\sigma_t \mu )^2} \big[
  ( 1 +   \sigma_t^2 L^2  + 2\sigma_t^2 \E ( {\rm Lip}(C_t )^2 | \mathcal{F}_{t-1}) ) \| z_{t-1} -z_\ast\|^2
  +  2 \sigma_t^2 \E ( \| C_t (z_{\ast})\|^2 | \mathcal{F}_{t-1})
   \big].
\EEAS
By denoting $\eta_t = \frac{\sigma_t \mu }{1+\sigma_t \mu} \in [0,1)$, we get
\BEAS
\E  
\| z_t - z_\ast \|^2
& \leqslant  & \Big( 1 - 2 \eta_t + \eta_t^2 + 2 \eta_t^2 \frac{L^2}{\mu^2} 
 +  2 \eta_t^2 \frac{1}{\mu^2}    \E ( {\rm Lip}(C_t )^2 | \mathcal{F}_{t-1}) 
\Big) \| z_{t-1} -z_\ast\|^2
  + 2\frac{\eta_t^2}{\mu^2} \E ( \| C_t z_{\ast}\|^2 | \mathcal{F}_{t-1})
  \big].
\EEAS
By selecting $\eta_t = \frac{2}{(t+1) + 4\frac{L^2}{\mu^2} + \frac{4}{\mu^2}   \E ( {\rm Lip}(C_t )^2 | \mathcal{F}_{t-1}) } = \frac{2}{   t +1+ A}$, with $A =  4\frac{L^2}{\mu^2} + \frac{4}{\mu^2}   \E ( {\rm Lip}(C_t )^2 | \mathcal{F}_{t-1})$, we get:
\BEAS
\E  
\| z_t - z_\ast \|^2
& \leqslant &  ( 1 -   \eta_t ) \E \| z_{t-1} -z_\ast\|^2
  + 2\frac{\eta_t^2}{\mu^2} \E ( \| C_t z_{\ast}\|^2 )
  \big] \\
  & = &  \frac{t-1 + A}{t+1+A}\E \| z_{t-1} -z_\ast\|^2
  + \frac{8}{(t+1+A)^2}\frac{1}{\mu^2} \E ( \| C_t z_{\ast}\|^2 ) \\
  & \leqslant & \frac{A(1+A)}{(t+1+A)(t+A)} \|z_0 - z_\ast\|^2 
  + \frac{8}{\mu^2} \sum_{u=1}^t  \frac{(u+A)(u+1+A)}{(t+1+A)(t+A)}   \frac{1}{(u+1+A)^2} \E ( \| C_u z_{\ast}\|^2 ) \\
  & & \hspace*{5cm} \mbox{ by expanding the recursion } t \mbox{ times}, \\
  & \leqslant & \frac{A(1+A)}{(t+1+A)(t+A)}  \|z_0 - z_\ast\|^2 
  + \frac{8}{\mu^2} \sum_{u=1}^t  \frac{1}{(t+1+A)(t+A)}   \E ( \| C_u z_{\ast}\|^2 )
\\
 & \leqslant & \frac{(1+A)^2}{ (t+A)^2}  \|z_0 - z_\ast\|^2 
  + \frac{8}{\mu^2 ( t + A) }  \sup_{u \in \{1,\dots,t\}}   \E ( \| C_u z_{\ast}\|^2 ).
\EEAS
The overall convergence rate is in $O(1/t)$ and the constant depends on the noise in the operator values at the optimum. Note that initial conditions are forgotten at a rate $O(1/t^2)$.

\textbf{Application to sampling from a finite family.}
When sampling from $|\I|$ operators $B_i$, $i \in \I$, and selecting $i_t$ with probability vector $\pi$, then
we have $ \E ( {\rm Lip}(C_t )^2 | \mathcal{F}_{t-1}) \leqslant   \bar{L}(\pi)^2 = \bar{L}^2$ defined as
$\sup_{\| z - z'\| \leqslant 1} \sqrt{ \sum_{i \in \I} \frac{1}{\pi_i} \| B_i(z) - B_i(z') \|^2 }$.
Thus, we can  take the step-size
$
\frac{2}{t+1 + 4 \frac{L^2 + \bar{L}^2}{\mu^2} }
$, 
which leads to $\sigma_t = \frac{ 2 / \mu}{t + 1 + 4 \frac{L^2 + \bar{L}^2}{\mu^2} }$. Moreover, if $L$ is unknown (or hard to compute), we can take $\bar{L}$ instead.

We may further bound: $
 \E ( \| C_u z_{\ast}\|^2 ) \leqslant 2  \E ( \| C_u z_{0}\|^2 ) +  2\E ( {\rm Lip}(C_t )^2 ) \| z_0 - z_\ast\|^2
$, and thus, if we start from an initial point $z_0$ such that $C_u z_{0}=0$, which is always possible for bi-linear problems, we get an overall bound of (taking $L = \bar{L}$ for simplicity)
\[
\Big( \frac{(1+ 8 \bar{L}^2 / \mu^2 )^2}{ (t+8 \bar{L}^2 / \mu^2)^2}    + \frac{16 \bar{L}^2 / \mu^2 }{ t + 8 \bar{L}^2 / \mu^2  }  \Big) \|z_0 - z_\ast\|^2 
 \leqslant  \frac{1 + 24 \bar{L}^2 / \mu^2 }{ t + 8 \bar{L}^2 / \mu^2  }   \|z_0 - z_\ast\|^2 .
\]
We thus get an overall $O(1/t)$ convergence rate.

\section{Proof for New Stochastic Algorithms}

 \vspace*{-.111cm}

We also consider the monotone operator set-up, since this is the only assumption that we use. We follow the proof of the corresponding convex minimization algorithms, with key differences which we highlight below. In particular, (a) we do not use function values, and (b) we use shorter step-sizes to tackle the lack of co-coercivity.

\subsection{SVRG: Stochastic-Variance reduced saddle-point problems (Theorem~\ref{theo:svrg})}

\vspace*{-.051cm}

\label{app:svrg}
We only analyze a single epoch starting from the reference estimate $\tilde{z}$, and show that the expected squared distance to optimum is shrunk by a factor of $3/4$ if the number of iterations per epoch is well-chosen. The epoch is started with $z_0 = \tilde{z}$.

 We denote by $\mathcal{F}_{t-1}$ the information up to time $t-1$. We consider sampling
$
i_{t1},\dots,i_{tm} \in \I
$ with replacement at time $t$. By using the contractivity of the resolvent operator of $A$, and the fact that $z_\ast = ( \idm + \sigma A)^{-1}( z_\ast - \sigma B(z_\ast) )$, we get:
\BEAS
\| z_t - z_\ast \|^2 & \leqslant & 
 \frac{1}{(1+\sigma \mu)^2} \Big\| z_{t-1} -z_\ast -   
 \sigma [ B (\tilde{z}) - B (z_\ast) + \frac{1}{m}\sum_{k=1}^m \frac{1} {\pi_{i_{tk}}}( B_{i_{tk}} (z_{t-1}) -  B_{i_{tk}} (\tilde{z}))] \Big\|^2\\
 & = & 
 \frac{1}{(1+\sigma \mu)^2} \Big\| z_{t-1} -z_\ast\\
 & &    -   
 \sigma [ B (z_{t-1}) - B (z_\ast)   +  \frac{1}{m}\sum_{k=1}^m \frac{1} {\pi_{i_{tk}}}( B_{i_{tk}} (z_{t-1}) -  B_{i_{tk}} (\tilde{z})) - (B(z_{t-1})  - B(\tilde{z})) ]\Big\|^2.
 \EEAS
 Expanding the squared norm, taking conditional expectations with $\E  ( \frac{1} {\pi_{i_{tk}}} B_{i_tk} | \mathcal{F}_{t-1}) = B$, and using the independence of $i_{t1},\dots,i_{tm}$, we get:
 \BEAS
 & &   \E \big[ \| z_t - z_\ast \|^2  | \mathcal{F}_{t-1} \big] \\
 & \leqslant & \frac{1}{(1+\sigma \mu)^2}  \big( \| z_{t-1} -z_\ast\|^2 
  - 2
\sigma (z_{t-1} -z_\ast)^\top  ( B (z_{t-1}) - B (z_\ast))  + \sigma^2 \| B (z_{t-1}) - B (z_\ast)\|^2
\big) \\
& & \hspace*{1cm} +   \frac{1}{m} \E \Big[
 \frac{1}{(1+\sigma \mu)^2} \Big\|
  \frac{1} {\pi_{i_t}}( B_{i_t} (z_{t-1}) -  B_{i_t} (\tilde{z})) - (B(z_{t-1})  - B(\tilde{z}))
 \Big\|^2 \Big|  \mathcal{F}_{t-1} 
\Big].
 \EEAS
 Using the monotonicity of $B$ and the Lipschitz-continuity of $B$  (like in Appendix~\ref{app:fb}) , we get the bound
 \[
     \frac{1+ \sigma^2 L^2}{(1+\sigma \mu)^2} \| z_{t-1} -z_\ast \|^2 + \frac{1}{m} \E \Big[
 \frac{1}{(1+\sigma \mu)^2} \big\|
  \frac{1} {\pi_{i_t}}( B_{i_t} (z_{t-1}) -  B_{i_t} (\tilde{z})) - (B(z_{t-1})  - B(\tilde{z}))
 \big\|^2 \big|  \mathcal{F}_{t-1} 
\Big].\]
We denote by $\bar{L}^2$ the quantity $\bar{L}^2 = \sup_{z,z' \in \mathcal{E}} \frac{ 1}{\| z - z'\|^2}\sum_{i \in \I} \frac{1}{\pi_i} \| B_i (z) - B_i (z') \|^2$. We then have  (using the fact that a variance is less than the second-order moment):
\[
 \E \Big[
  \big\|
  \frac{1} {\pi_{i_t}}( B_{i_t} (z_{t-1}) -  B_{i_t} (\tilde{z})) - (B(z_{t-1})  - B(\tilde{z}))
 \big\|^2 \big|  \mathcal{F}_{t-1} 
\Big] \leqslant 
 \E \Big[
  \big\|
  \frac{1} {\pi_{i_t}}( B_{i_t} (z_{t-1}) -  B_{i_t} (\tilde{z})) 
 \big\|^2 \big|  \mathcal{F}_{t-1} 
\Big],
\]
which is less than $\bar{L}^2 \| z_{t-1} - \tilde{z} \|^2$ because we sample $i_t$ from $\pi$.
 This leads to 
\BEAS
 \E \big[ \| z_t - z_\ast \|^2  | \mathcal{F}_{t-1} \big] & \leqslant & \frac{1+ \sigma^2 L^2}{(1+\sigma \mu)^2} \| z_{t-1} -z_\ast \|^2 + 
 \frac{1}{(1+\sigma \mu)^2} \frac{\bar{L}^2}{m} \| z_{t-1} - \tilde{z} \|^2 
 \\
& \leqslant & \Big( 1 - 2 \eta + \eta^2 + \eta^2 \frac{L^2}{\mu^2}  + \frac{(1+a) \eta^2}{\mu^2} \frac{\bar{L}^2}{m} \Big)\| z_{t-1} -z_\ast \|^2  \\
& & \hspace*{5cm}
+ \frac{(1+a^{-1}) \eta^2}{\mu^2} \frac{\bar{L}^2}{m} \|   \tilde{z}  -z_\ast \|^2 ,
\EEAS
with $\eta = \frac{\sigma \mu }{1+\sigma \mu} \in [0,1)$ and $a>0$ to be determined later.
Assuming that $ \eta  \Big( 1 +    \frac{L^2}{\mu^2}  + \frac{(1+a)   }{\mu^2} \frac{\bar{L}^2}{m}  \Big) \leqslant 1$, and taking full expectations, this leads to:
\[ \E  \|z_t - z_\ast\|^2 \leqslant ( 1- \eta) \E  \|z_{t-1} - z_\ast\|^2 + \frac{(1+a^{-1}) \eta^2}{\mu^2} \frac{\bar{L}^2}{m} \|   \tilde{z} -z_\ast \|^2 ,
\]
that is we get a shrinking of the expected distance to optimum with additional noise that depends on the distance to optimum of the reference point $\tilde{z}$. The difference with the convex minimization set-up of~\cite{xiao2014proximal} is that the proof is more direct, and we get a shrinkage directly on the iterates (we have no choice for monotone operators), without the need to do averaging of the iterates. Moreover, we never use any monotonicity of the operators $B_i$, thus allowing any type of splits (as long as the sum $B$ is monotone).

Then, using the fact that $z_0 =\tilde{z}$, and expanding the recursion:
\BEAS
\E  \|z_t - z_\ast\|^2 & \leqslant & 
( 1- \eta)^t \| z_0 -z_\ast\|^2 + \big( \sum_{u=0}^{t-1} (1-\eta)^u \big)  \frac{(1+a^{-1}) \eta^2}{\mu^2} \frac{\bar{L}^2}{m} \|   \tilde{z} -z_\ast \|^2 \\
& \leqslant & 
\bigg(
 ( 1 - \eta  )^t + 
 {\frac{(1+a^{-1})  \eta }{\mu^2} \frac{\bar{L}^2}{m} } \bigg)\|   \tilde{z} -z_\ast \|^2 .
\EEAS
If we take $a=2$,   $\eta = \frac{1}{ \textcolor{white}{\big|} 1+  L^2 +  3 \bar{L}^2 / ( m \mu^2 )}$, which corresponds to
$\sigma = \frac{1}{\mu} \frac{ \eta}{1-\eta} = \frac{\mu}{\textcolor{white}{\big|} L^2 + \frac{3}{m} \bar{L}^2}$
and $t = \log 4 / \eta = \log 4 \times ( 1 +  \frac{L^2}{\mu^2} + 3 \frac{\bar{L}^2}{m\mu^2})$, we obtain  a bound of $3/4$, that is, after $t$ steps in an epoch, we obtain 
$\E  \|z_t - z_\ast\|^2  \leqslant \frac{3}{4} \|   \tilde{z} -z_\ast \|^2$, which is the desired result.

In terms of running-time, we therefore need a time proportional to $T(B) + \big( 1 +  \frac{L^2}{\mu^2} + 3 \frac{\bar{L}^2}{m\mu^2} \big)
\big(  m \max_{i}  T(B_i)  + T_{\rm prox}(A) \big)$,  times $\log \frac{1}{\varepsilon}$ to reach precision $\varepsilon$.

 Note that if $L^2$ is too expensive to compute (because it is a global constant), we may replace it by $\bar{L}^2$ and get a worse bound (but still a valid algorithm).

\subsection{SAGA: Online stochastic-variance reduced saddle-point problems (Theorem~\ref{theo:saga})}

\vspace*{-.051cm}

\label{app:saga}
The proof follows closely the one of SVRG above. Following the same arguments, we get, by contractivity of the resolvent operator:
\BEAS
\| z_t - z_\ast \|^2 & \leqslant & 
 \frac{1}{(1+\sigma \mu)^2} \Big\| z_{t-1} -z_\ast -   
 \sigma [  \sum_{i \in \I} g_{t-1}^{i} - B(z_\ast) + \frac{1}{m} \sum_{k=1}^m \frac{1} {\pi_{i_{tk}}}(   B_{i_{tk}} (z_{t-1}) - g_{t-1}^{i_{tk}} )] \Big\|^2\\
 & = & 
 \frac{1}{(1+\sigma \mu)^2} \Big\| z_{t-1} -z_\ast -   
 \sigma [ B (z_{t-1}) - B(z_\ast)  \\
  & & \hspace*{2cm} + \frac{1}{m} \sum_{k=1}^m  \frac{1} {\pi_{i_{tk}}}( B_{i_{tk}} (z_{t-1}) -   g_{t-1}^{i_{tk}}) - (B(z_{t-1})  - \sum_{i \in \I} g_{t-1}^{i}) ]\Big\|^2.
  \EEAS
  Then, using independence, monotonicity and Lipschitz-continuity of $B$, we get (note that we never use any monotonicity of $B_i$), like in the proof of Theorem~\ref{theo:svrg}:
  \BEAS
\E \big[ \| z_t - z_\ast \|^2  | \mathcal{F}_{t-1} \big] & \leqslant & \frac{1+ \sigma^2 L^2}{(1+\sigma \mu)^2} \| z_{t-1} -z_\ast \|^2  \\
& & \hspace*{0cm} + \frac{1}{m}\E \Big[
 \frac{1}{(1+\sigma \mu)^2} \big\|
  \frac{1} {\pi_{i_t}}( B_{i_t} (z_{t-1}) -      g_{t-1}^{i_t}) - (B(z_{t-1})  -\sum_{i \in \I} g_{t-1}^{i})
 \big\|^2 \big|  \mathcal{F}_{t-1} 
\Big] \\
& \leqslant & \frac{1+ \sigma^2 L^2}{(1+\sigma \mu)^2} \| z_{t-1} -z_\ast \|^2 + \frac{1}{m}
 \frac{1}{(1+\sigma \mu)^2}  \Big( \sum_{i \in \I} \frac{1}{\pi_i}   \| B_i (z_{t-1} ) -  g_{t-1}^i\|^2 \Big)
 \\
 & \leqslant & \Big( 1 - 2 \eta + \eta^2 + \eta^2 \frac{L^2}{\mu^2}  + \frac{(1+a) \eta^2}{\mu^2} \frac{ \bar{L}^2}{m} \Big)\| z_{t-1} -z_\ast \|^2  \\
 & & \hspace*{4cm}
+ \frac{(1+a^{-1}) \eta^2}{\mu^2 m} \Big( \sum_{i \in \I} \frac{1}{\pi_i}   \| B_i (z_{\ast})  -  g_{t-1}^i\|^2 \Big),
 \EEAS
 with $\eta = \frac{ \sigma \mu}{ 1 + \sigma \mu}$.
 Assuming $\eta \big( 1 +  \frac{L^2}{\mu^2}  + \frac{(1+a)  }{\mu^2} \frac{ \bar{L}^2}{m} \big) \leqslant 1$, we get
 \[
 \E \big[ \| z_t - z_\ast \|^2  | \mathcal{F}_{t-1} \big]
  \leqslant (1 - \eta ) \| z_{t-1} -z_\ast \|^2+ \frac{(1+a^{-1}) \eta^2}{\mu^2 m} \Big( \sum_{i \in \I} \frac{1}{\pi_i}   \| B_i (z_{\ast})  -  g_{t-1}^i\|^2 \Big).
 \]
 Like in the SVRG proof above, we get a contraction of the distance to optimum, with now an added noise that depends on the difference between our stored operator values and the operator values at the global optimum. We thus need to control this distance by adding the proper factors to a Lyapunov function. Note that we never use any monotonicity of the operators $B_i$, thus allowing any type of splits (as long as the sum $B$ is monotone).
 
We assume that we update (at most $m$ because we are sampling with replacement and we may sample the same gradient twice) ``gradients'' $g_{t}^i$ uniformly at random (when we consider uniform sampling, we can reuse the same gradients as dependence does not impact the bound), by replacing them by
$
g_t^i = B_i (z_{t-1})
$.  
Thus:
\BEAS
& & \E \Big( \sum_{i \in \I} \frac{1}{\pi_i}   \| B_i (z_{\ast} ) -  g_{t}^i\|^2   \Big| \mathcal{F}_{t-1} \Big) \\
&  =  &  \E \Big( 
\sum_{i \ \rm selected} \frac{1}{\pi_i}   \| B_i (z_{\ast} ) -  B_i (z_{t-1})\|^2
+  \sum_{i \ \rm non \ selected} \frac{1}{\pi_i}   \| B_i (z_{\ast} )   -  g_{t-1}^i\|^2  \Big| \mathcal{F}_{t-1} \Big)  \\
&  =  & \!\!\!\!\!
 \E \Big( \sum_{i \ \rm selected} \frac{1}{\pi_i} \Big(   \| B_i (z_{\ast} ) -  B_i (z_{t-1})\|^2 -  \| B_i (z_{\ast} )   -  g_{t-1}^i\|^2 \Big)
+  \sum_{i \in \I } \frac{1}{\pi_i}   \| B_i (z_{\ast} )   -  g_{t-1}^i\|^2 \Big| \mathcal{F}_{t-1} \Big).
\EEAS
Since we sample uniformly \emph{with} replacement, the marginal probabilities of selecting an element $i$
is equal to $\rho = 1 - ( 1 - \frac{1}{|\I|} )^m$.
We thus get
\BEAS
\E \Big( \sum_{i \in \I} \frac{1}{\pi_i}   \| B_i (z_{\ast} ) -  g_{t}^i\|^2   \Big| \mathcal{F}_{t-1} \Big)
& \leqslant &  ( 1 - \rho)  \sum_{i \in \I } \frac{1}{\pi_i}   \| B_i (z_{\ast} )   -  g_{t-1}^i\|^2 
+ \rho   \sum_{ i \in \I} \frac{1}{\pi_i}   \| B_i (z_{\ast} ) -  B_i (z_{t-1})\|^2
\\
& \leqslant &  ( 1 - \rho)  \sum_{i \in \I } \frac{1}{\pi_i}   \| B_i (z_{\ast} )   -  g_{t-1}^i\|^2 
+ \rho  \bar{L}^2  \| z_{t-1} - z_\ast\|^2.
 \EEAS
Therefore, overall, we have, for  a scalar $b>0$ to be chosen later:
\BEAS
& & \E \Big(\| z_t - z_\ast \|^2  +  b \sum_{i \in \I} \frac{1}{\pi_i}   \| B_i (z_{\ast} ) -  g_{t}^i\|^2   \Big| \mathcal{F}_{t-1} \Big) \\
& \leqslant &  \Big( 1 - 2 \eta + \eta^2 + \eta^2 \frac{L^2}{\mu^2}  + \frac{(1+a) \eta^2}{\mu^2}  \frac{\bar{L}^2}{m} +  b \rho \bar{L}^2   \Big) \| z_{t-1} - z_\ast \|^2  \\
& & \hspace*{3cm} +   b \Big( 1 - \rho +  b^{-1} \frac{(1+a^{-1}) \eta^2}{ m \mu^2} \Big)  \sum_{i \in \I } \frac{1}{\pi_i}   \| B_i (z_{\ast} )   -  g_{t-1}^i\|^2 .
\EEAS
If we take $a=2$,  $\eta =   \frac{1}{ \max \{ \frac{3|\I|}{ 2m}, 1 +  \frac{L^2}{\mu^2} + 3 \frac{\bar{L}^2}{ m  \mu^2} \} }$, 
which corresponds to
$\sigma = \frac{1}{\mu} \frac{ \eta}{1-\eta} = \frac{\mu}{\textcolor{white}{\big|} 
\max \{ \frac{3|\I|}{2 m} -1 ,  \frac{L^2}{\mu^2} + 3 \frac{\bar{L}^2}{ m  \mu^2} \} }$,
with $b \rho   \bar{L}^2   = \frac{3 \eta}{4}$, then we get the bound (using $ \eta \leqslant 1/ ( \bar{L}^2/(3m) )$):
\[
(1 - \frac{\eta}{4}) \| z_{t-1} - z_\ast \|^2 + ( 1 - \frac{\rho}{3} )  \sum_{i \in \I } \frac{1}{\pi_i}   \| B_i (z_{\ast} )   -  g_{t-1}^i\|^2 
,\]
which shows that the function $ (z,g)\mapsto \| z - z_\ast \|^2  +  b \sum_{i \in \I} \frac{1}{\pi_i}   \| B_i (z_{\ast} ) -  g^i\|^2 $ is a good Lyapunov function for the problem that shrinks geometrically in expectation (it resembles the one from convex minimization, but without the need for function values).

Finally, since we assume that $m \leqslant |\I|$, we have
$\rho  = 1  - ( 1- 1/|\I|)^m \geqslant 1 - \exp( - m / |\I| ) \geqslant m / (2|\I| )$.
This leads to, after $t$ iterations
\[
\E \|z_t - z_\ast\|^2 \leqslant  ( 1 - \min\{ \frac{\eta}{4},  \frac{m}{6|\I|} \}  )^t  
\Big[ \| z_0 - z_\ast\|^2  + \frac{3 \eta}{4 \rho \bar{L}^2}
\sum_{i \in \I} \frac{1}{\pi_i}   \| B_i (z_{\ast} ) -  B_i(z_0)\|^2 \Big].
\]
We have $\eta \leqslant 2 m / ( 3|\I| )$ and $ 3 \eta  / (4 \rho) \leqslant \frac{3}{4} \frac{2m}{3|\I|} \frac{2|\I|}{m} \leqslant 1$, leading to 
\[
\E \|z_t - z_\ast\|^2 \leqslant  2 ( 1 -  \frac{\eta}{4}  )^t  
  \| z_0 - z_\ast\|^2,  \]
which is the desired result.

Note that we get the same overall running-time complexity than for SVRG.

\textbf{Factored splits.}
Note that when applying to saddle-points with factored splits, we need to use a Lyapunov function that considers these splits. The only difference is to treat separately the two parts of the vectors, leading to replacing everywhere $|\I|$ by $\max \{|\mathcal{J}|, |\mathcal{K}|\}$. 

\subsection{Acceleration}
\label{app:accsvrg}

\vspace*{-.051cm}

We also consider in this section a proof based on monotone operators. We first give the algorithm for saddle-point problems.

\textbf{Algorithms for saddle-point problems.}
At each iteration, we need solve the problem in \eq{ppa} of the main paper,  with the SVRG algorithm applied to
$\tilde{K}(x,y) = K(x,y) - \lambda \tau x^\top \bar{x} + \gamma \tau y^\top \bar{y}$,  and $\tilde{M}(x,y) = M(x,y) + \frac{\lambda \tau}{2} \| x\|^2 - \frac{\gamma \tau}{2} \| y \|^2$. These functions lead to constants $\tilde\lambda= \lambda ( 1+ \tau)$, $\tilde\gamma = \gamma (1 + \tau)$ and $\tilde L = L / ( 1+ \tau)$, $\tilde \sigma  = \sigma ( 1 + \tau)^2$.
We thus get the iteration, for a single selected operator,
\[ (x,y) \leftarrow  {\rm prox}_{\tilde{M}}^{\tilde{\sigma}} \big[ (x,y) - \tilde{\sigma} \DDtilde \big( \tilde{B}(\tilde{x},\tilde{y}) +    \big\{ \frac{1}{\pi_{i}} \tilde{B}_{i}(x,y) - \frac{1}{\pi_{i}} \tilde{B}_i(\tilde{x},\tilde{y}) \big\}\big)\big].
\]
A short calculation shows that 
$ {\rm prox}_{\tilde{M}}^{\tilde{\sigma}} (x,y) = 
 {\rm prox}_{M}^{ \sigma(1+\tau) / ( 1 + \sigma \tau ( 1 + \tau) )} ( (x,y) / ( 1 + \sigma \tau ( 1+ \tau) ) )$, leading to the update (with $\sigma$ the step-size from the regular SVRG algorithm in \mysec{svrg}):
 \[
\!\!\!\!\!\! (x,y) \leftarrow    {\rm prox}_{\tilde{M}}^{\tilde{\sigma}}   \big[ (x,y) 
 + \sigma \tau (1 + \tau) ( \bar{x},\bar{y}) -  {\sigma ( 1 + \tau) } \DD \big(  {B}(\tilde{x},\tilde{y}) +    \big\{ \frac{1}{\pi_{i}} \tilde{B}_{i}(x,y) - \frac{1}{\pi_{i}} \tilde{B}_i(\tilde{x},\tilde{y}) \big\}\big)\big].
\]
This leads to Algorithm~\ref{algo:svrg-acc}, where differences with the SVRG algorithm, e.g., Algorithm~\ref{algo:svrg}, are highlighted in red. Given the value of $\tau$, the estimate $(\bar{x},\bar{y})$ is updated every $\log(1+\tau)$ epochs of SVRG. While this leads to a provably better convergence rate, in practice, this causes the algorithm to waste time solving with too high precision the modified problem. We have used the simple heuristic of changing $(\bar{x},\bar{y})$ one epoch after  the primal-dual gap has been reduced from the previous change of $(\bar{x},\bar{y})$.


\begin{algorithm}
\caption{Accelerated Stochastic Variance Reduction for Saddle Points} \label{algo:svrg-acc}
\begin{algorithmic}

\REQUIRE Functions $(K_i)_i$, $M$, probabilities $(\pi_i)_i$, smoothness $\bar{L}(\pi)$ and $L$, iterate $(x,y)$\\
\hspace*{.69cm} number of epochs $v$, number of updates per iteration $m$, \textcolor{red}{acceleration factor $\tau$}
\STATE Set $\sigma=  \big[  L^2 +  3\bar{L}^2/m\big]^{-1}$  and \textcolor{red}{$(\bar{x},\bar{y}) = (x,y)$}
\FOR{$u$ = $1$ to $v$ }
\STATE \textcolor{red}{If $u = 0 \mod \lceil 2 +  2\log( 1 + \tau)/(\log 4/3) \rceil$, set $(\bar{x},\bar{y}) = (\tilde{x},\tilde{y})$}
\STATE Initialize $(\tilde{x},\tilde{y}) = (x,y)$ and compute $B(\tilde{x},\tilde{y})$
    \FOR{$k$ = $1$ to $\log 4 \times ( L^2 +   3   \bar{L}^2 / m )\textcolor{red}{ ( 1 +\tau)^2} $ }
    \STATE Sample $i_1,\dots,i_m \in \I$ from probability vector $(\pi_i)_i$ with replacement
    \STATE   $z \leftarrow (x,y) 
 \textcolor{red}{+ \sigma \tau (1 + \tau) ( \bar{x},\bar{y})}  -  {\sigma \textcolor{red}{( 1 + \tau)} } \DD \big(  {B}(\tilde{x},\tilde{y}) +    \big\{ \frac{1}{\pi_{i}} \tilde{B}_{i}(x,y) - \frac{1}{\pi_{i}} \tilde{B}_i(\tilde{x},\tilde{y}) \big\}\big)$
\STATE $(x,y) \leftarrow  {\rm prox}_{M}^{ \sigma\textcolor{red}{(1+\tau) / ( 1 + \sigma \tau ( 1 + \tau) )}} ( z  \textcolor{red}{/ ( 1 + \sigma \tau ( 1+ \tau) ) })$
    \ENDFOR
\ENDFOR
\ENSURE Approximate solution $(x,y)$
\end{algorithmic}
\end{algorithm}

\textbf{Proof of Theorem~\ref{theo:acc} using monotone operators.}
We consider $\tau \geqslant 0$, and we consider the following algorithm, which is the transposition of the algorithm presented above. We consider a mini-batch  $m=1$ for simplicity. We consider a set of SVRG epochs, where $\bar{z}$ remains fixed. These epochs are initialized by $\tilde{z} = \bar{z}$.

For each SVRG epoch, given $\bar{z}$ and $\tilde{z}$, and starting from $z = \tilde{z}$, we run $t$ iterations of:
\[
z \leftarrow 
 (   \idm + \sigma ( \tau \idm +  A) )^{-1} \big( z  - \sigma [ B \tilde{z} + \frac{1} {\pi_{i}}( B_{i} z -  B_{i} \tilde{z} )
 - \tau \bar{z}
 ] \big),
\]
and then update $\tilde{z}$ as $z$ at the end of the SVRG epoch.
It corresponds exactly to running the SVRG algorithm   to find $ (\tau \idm + A + B)^{-1} ( \tau \bar{z})$ approximately,
we know from the proof of Theorem~\ref{theo:svrg} that after   $ \log 4 \big( 1 + \frac{L^2}{\mu^2 ( 1 + \tau)^2} +  \frac{L^2}{\mu^2 ( 1 + \tau)^2}\big)$ iterations, we have an iterate $z$ such that
$\E \| z - ( \tau \idm + A + B )^{-1} (\tau \bar{z})\|^2 \leqslant \frac{3}{4} \E \| \tilde{z} - ( \tau \idm + A + B )^{-1} (\tau \bar{z})\|^2$. Thus, if we run $s$ epochs where we update~$\tilde{z}$ (but not $\bar{z}$) at each start of epoch, we get an iterate $z$ such that
$\E \| z - ( \tau \idm + A + B )^{-1} (\tau \bar{z})\|^2 \leqslant (\frac{3}{4})^s \E \| \bar{z} - ( \tau \idm + A + B )^{-1} (\tau \bar{z})\|^2$, and thus
\BEAS
& &  \E \| z - ( \tau \idm + A + B )^{-1} (\tau \bar{z})\|^2  \\
& \leqslant  & \Big( \frac{3}{4} \Big)^s \E \| \bar{z} - ( \tau \idm + A + B )^{-1} (\tau \bar{z})\|^2  \\
& =   & \Big( \frac{3}{4} \Big)^s \E\| \bar{z} - z_\ast - ( \tau \idm + A + B )^{-1} (\tau \bar{z}) +  ( \tau \idm + A + B )^{-1} (\tau z_\ast)\|^2  \\
 & & \hspace*{3cm} \mbox{using } z_\ast = ( \tau \idm + A + B )^{-1} (\tau z_\ast), \\
 & = & \Big( \frac{3}{4} \Big)^s \E\| \bar{z} - z_\ast - (   \idm + \tau^{-1}(A + B) )^{-1} (  \bar{z}) +  (   \idm + \tau^{-1}(A + B) )^{-1} (  z_\ast)\|^2 .
 \EEAS
 We may now use the fact that for any multi-valued maximal monotone operator $C$, $\idm - ( I + C)^{-1} = (\idm + C^{-1})^{-1}$, which shows that $\idm - ( I + C)^{-1} $ is $1$-Lipschitz-continuous. Thus, after $s$ epochs of SVRG,
 \BEAS
\E\| z - ( \tau \idm + A + B )^{-1} (\tau \bar{z})\|^2  & \leqslant   & \Big( \frac{3}{4} \Big)^s \E\|\bar{z} - z_\ast\|^2.  
\EEAS
This implies, by Minkowski's inequality, 
\BEAS
 & & (\E\| z - z_\ast \|^2)^{1/2} \\
 & \leqslant & (\E\| z - ( \tau \idm + A + B )^{-1} (\tau \bar{z}) \|^2)^{1/2}+ (\E \| ( \tau \idm + A + B )^{-1} (\tau \bar{z})- z_\ast \|^2)^{1/2} \\
& \leqslant  & \Big( \frac{3}{4} \Big)^{s/2} (\E \|\bar{z} - z_\ast\|^2)^{1/2} +  (\E \| ( \tau \idm + A + B )^{-1} (\tau \bar{z})- ( \tau \idm + A + B )^{-1} (\tau z_\ast) \|^2)^{1/2}  \\
& =  & \Big( \frac{3}{4} \Big)^{s/2} (\E \|\bar{z} - z_\ast\|^2)^{1/2} +  (\E \| (   \idm + \tau^{-1}(A + B) )^{-1} (  \bar{z})-(   \idm + \tau^{-1}(A + B) )^{-1} (  z_\ast) \|^2)^{1/2}  \\
& \leqslant  & \Big( \frac{3}{4} \Big)^{s/2} (\E\|\bar{z} - z_\ast\|^2)^{1/2} +\frac{1}{1 +\tau^{-1} \mu} (\E \|\bar{z} - z_\ast\|^2)^{1/2} \\
& = &  \Big( \frac{3}{4} \Big)^{s/2} (\E \|\bar{z} - z_\ast\|^2)^{1/2} +\frac{\tau }{ \tau +  \mu} (\E\|\bar{z} - z_\ast\|^2)^{1/2},
\EEAS
using the fact that the contractivity of resolvents of strongly monotone operators.
Thus after $s = 2 + 2 \frac{ \log ( 1 + \frac{\tau}{\mu} ) }{\log \frac{4}{3}}$, we get a decrease by
$( 1 - \frac{\mu}{\tau + \mu})$, and thus the desired result.

\subsection{Factored splits and bi-linear models}

\vspace*{-.051cm}

\label{app:factored}
In the table below, we report the running-time complexity for the factored splits which we used in simulations. Note that SAGA and SVRG then have different bounds. Moreover, all these schemes are adapted when $n$ is close to $d$. For $n$ much different from $d$, one could imagine 
to (a) either complete with zeros or (b) to regroup the data in the larger dimension so that we get as many blocks as for the lower dimension. 

\begin{table}[H]
\renewcommand{\tabcolsep}{0.13cm}
\begin{center}
\begin{tabular}{|l|rll|}
\hline
Algorithms & & Complexity &   \\
\hline
Stochastic FB-non-uniform  &  $ (1 / {\varepsilon})\ \times \big(\!$ &  $  \max \{n,d\}   \| K\|_F^2  / ( \lambda \gamma)   \textcolor{white}{\Big|}  $ & $\big)$   \\ 
Stochastic FB-uniform  &  $ (1 / {\varepsilon})\ \times \big(\!$ &  $  \textcolor{red}{\max \{n,d\} ^2} \| K\|_{\rm max}^2  / ( \lambda \gamma)  \textcolor{white}{\Big|}   $   & $\big)$  \\ 
\hline
 SVRG-uniform   &  $\log(1 / {\varepsilon})\ \times \big(\!$ &  $
nd +  \textcolor{red}{\max \{n,d\} ^2}  \|K\|_{\rm max}^2  / ( \lambda \gamma)  \textcolor{white}{\Big|}  \!\!\!\!
   $ & $\big)$ \\ 
SAGA-uniform   &  $\log(1 / {\varepsilon})\ \times \big(\!$ &  $
\textcolor{red}{\max \{n,d\} ^2} +  \textcolor{red}{\max \{n,d\} ^2}  \|K\|_{\rm max}^2  / ( \lambda \gamma)  \textcolor{white}{\Big|}  \!\!\!\!
   $ & $\big)$ \\ 
 SVRG-non-uniform   &  $\log(1 / {\varepsilon})\ \times \big(\!$ &  $
nd + \max \{n,d\} \|K\|_F^2  / ( \lambda \gamma)    \textcolor{white}{\Big|} 
  $ & $\big)$\\ 
SAGA-non-uniform   &  $\log(1 / {\varepsilon})\ \times \big(\!$ &  $
\textcolor{red}{\max \{n,d\} ^2}  + \max \{n,d\} \|K\|_F^2  / ( \lambda \gamma)    \textcolor{white}{\Big|} 
  $ & $\big)$\\ 
SVRG-non-uniform-acc.    &  $\log(1 / {\varepsilon})\ \times \big(\!$ &  $ nd + 
 \textcolor{red}{\max \{n,d\} ^{3/2}}  \| K \|_F / \sqrt{\lambda \gamma}\textcolor{white}{\Big|} 
    $ & $\big)$\\ 
\hline
\end{tabular}
\end{center}
\caption{Summary of convergence results for the strongly $(\lambda,\gamma)$-convex-concave bilinear saddle-point problem with matrix $K$ and factored splits, with access to a single row and a single column per iteration. The difference with the individual splits from Table~\ref{table:summary} is highlighted in red.  
  \label{table:summaryfacgored}}
\vspace*{-.5cm}
\end{table}
 
\section{Surrogate to Area Under the ROC Curve}
\label{app:auc}

 \vspace*{-.111cm}
 
We consider the following loss function on $\rb^n$, given a vector of positive and negative labels, which corresponds to a convex surrogate to the number of misclassified pairs~\cite{joachims2005support,herbrich1999large}:
\BEAS
\ell(u) &= &  \frac{1}{2 n_+ n_-} \sum_{ i_+ \in I_+} \sum_{i_- \in I_-} (  1 - u_{i_-} + u_{i_+})^2
\\
&= &  \frac{1}{2 n_+ n_-} \sum_{ i_+ \in I_+} \sum_{i_- \in I_-}    \big\{ 1 + u_{i_-}^2 + u_{i_+}^2 - 2 u_{i_-} +2 u_{i_+} - 2u_{i_-} u_{i_+} \big\}  \\
& = & \frac{1}{2} + \frac{1}{n_+} \sum_{i_+ \in I_+} u_{i_+} 
-  \frac{1}{n_-} \sum_{i_-\in I_-} u_{i_-}  + \frac{1}{2n_-} \sum_{i_- \in I_-} u_{i_-}^2
+   \frac{1}{2n_+} \sum_{i_+ \in I_+} u_{i_+}^2 - \frac{1}{n_+ n_-} \sum_{ i_+ \in I_+} \sum_{i_- \in I_-}
u_{i_-} u_{i_+}
\\
& = & \frac{1}{2} + \frac{1}{n_+} e_+^\top u - \frac{1}{n_-} e_-^\top u + \frac{1}{2} u^\top \Diag(\frac{1}{n_+} e_+ + \frac{1}{n_-} e_-) u - \frac{1}{2n_+ n_-} u^\top ( e_+ e_-^\top + e_- e_+^\top)u \\
& = & \frac{1}{2} -  a^\top u + \frac{1}{2} u^\top A u,
\EEAS
with $e_+ \in \rb^n$ the indicator vector of $I_+$ and $e_- \in \rb^n$ the indicator vector of $I_-$.
We have $A =  \Diag(\frac{1}{n_+} e_+ + \frac{1}{n_-} e_-) - \frac{1}{n_+ n_-} \big[ e_+ e_-^\top + e_- e_+^\top \big]$
and $a = e_+ / n_+ - e_- / n_-$. A short calculation shows that the largest eigenvalue of $A$ is
$\frac{1}{M} = \frac{1}{n_+} + \frac{1}{n_-}$.

We consider the function $h(u) = \frac{1}{2} u^\top A u$. It is $(1/M)$-smooth, its Fenchel conjugate is equal to
\[
\frac{1}{2} v^\top A^{-1} v,
\]
and our function $g$ will be equal to $v \mapsto \frac{1}{2} v^\top A^{-1} v - \frac{M}{2 } \| v\|^2$. Given that $1$ is a singular vector of $A$, $g(v)$ is finite only when $v^\top 1_n = 0$.

We need to be able to compute $g(v)$, i.e., solve the system $A^{-1} v$, and to compute the the proximal operator
\[
\min_v \frac{1}{2} \| v - v_0\|^2 + \sigma g(v) = 
\min_v \frac{1}{2} \| v - v_0\|^2 + \frac{\sigma}{2} v^\top ( A^{-1} - M \idm ) v,
\]
which leads to to the system: $(A^{-1} - M \idm + \sigma^{-1}\idm )  v= \sigma^{-1} v_0$,
which is equivalent to:
$(\idm - M A + \sigma^{-1}A )  v= \sigma^{-1} A v_0$
We thus need to compute efficiently $ A w$, and $(\idm + \kappa A)^{-1} w$ with $\kappa > -M$.
We have
\BEAS
\idm + \kappa A
& = & \Diag( ( 1 + \kappa/n_+) e_+ + ( 1+ \kappa/n_-)e_- ) 
- \frac{\kappa}{n_+ n_-} \big[ e_+ e_-^\top + e_- e_+^\top \big] \\
& = & \Diag( ( 1 + \kappa/n_+) e_+   + ( 1+ \kappa/n_-)e_- ) ^{1/2} \\
& & 
\Big[
\idm - \frac{\kappa}{n_+ n_-} \big(
\big[\frac{1}{\sqrt{1+\kappa/n_+}} e_+ \big]
\big[\frac{1}{\sqrt{1+\kappa/n_-}} e_- \big]^\top  - 
\big[\frac{1}{\sqrt{1+\kappa/n_-}} e_- \big] \big[\frac{1}{\sqrt{1+\kappa/n_+}} e_+ \big]
^\top
\big) \Big]\\
& & 
 \Diag( ( 1 + \kappa/n_+) e_+ + ( 1+ \kappa/n_-)e_- ) ^{1/2} \\
&  = &  D^{1/2} ( \idm - \alpha u_+ u_-^\top - \alpha u_- u_+^\top ) D^{1/2},
\EEAS
with $u_+^\top u_- =0$ and $u_+ = \frac{e_+}{\sqrt{n_+}}, u_- = \frac{e_-}{\sqrt{n_-}}$ of norm 1
and $D = \Diag( ( 1 + \kappa/n_+) e_+ + ( 1+ \kappa/n_-)e_- )$.
We have:
\BEAS
  \idm - \alpha u_+ u_-^\top - \alpha u_- u_+^\top 
& = &  \idm -   u_+ u_+^\top -   u_- u_-^\top +
( u_+,  u_- ) 
\big(\!\!\! \begin{array}{cc} 1\! \!\!&\!\! -\alpha \\
 -\alpha \!\!&\!\!\! 1  \end{array}\!\!\big)
(u_+ , u_-)^\top
\\
 ( \idm - \alpha u_+ u_-^\top - \alpha u_- u_+^\top )^{-1}
& = &  \idm -   u_+ u_+^\top -   u_- u_-^\top +
\frac{1}{1-\alpha^2} ( u_+,  u_- ) 
\big(\!\!\! \begin{array}{cc} 1\! \!\!&\!\! \alpha \\
 \alpha \!\!&\!\!\! 1  \end{array}\!\!\big)
(u_+ , u_-)^\top \\
& = &  \idm + ( 1/(1-\alpha^2) - 1)    u_+ u_+^\top + ( 1/(1-\alpha^2) - 1)     u_- u_-^\top +
\frac{\alpha}{1-\alpha^2} ( u_+   u_-^\top + u_- u_+^\top ) 
\\
& = &  \idm + ( 1/(1-\alpha^2) - 1) \frac{1}{n_+}   e_+ e_+^\top + ( 1/(1-\alpha^2) - 1)    \frac{1}{n_-}   e_- e_-^\top   \\
& & +
\frac{\alpha}{1-\alpha^2} \frac{1}{\sqrt{n_+ n_-}} ( e_+   e_-^\top + e_- e_+^\top ) .
\EEAS
We have here $\alpha  = \frac{\kappa}{n_+ n_-} \sqrt{\frac{n_+}{1 + \kappa /n_+}}
\sqrt{\frac{n_-}{1 + \kappa /n_-}}$. Thus
\[(\idm + \kappa A)^{-1} = D^{-1/2} \big[
\idm -   u_+ u_+^\top -   u_- u_-^\top +
\frac{1}{1-\alpha^2} ( u_+,  u_- ) 
\big(\!\!\! \begin{array}{cc} 1\! \!\!&\!\! \alpha \\
 \alpha \!\!&\!\!\! 1  \end{array}\!\!\big)
(u_+ , u_-)^\top
\big] D^{-1/2},\]
which can be done in $O(n)$.

Moreover, we have
\BEAS
  A & = & \Diag( ( 1 /n_+) e_+   + ( 1 /n_-)e_- ) ^{1/2} \\
& & 
\Big[
\idm - \frac{1}{n_+ n_-} \big(
\big[ \sqrt{ n_+}  e_+ \big]
\big[\sqrt{ n_-}  e_- \big]^\top  - 
\big[\sqrt{ n_-}  e_-\big] \big[\sqrt{ n_+}  e_+\big]
^\top
\big) \Big]\\
& & 
  \Diag( ( 1 /n_+) e_+   + ( 1 /n_-)e_- ) ^{1/2} 
  \\
  &  = &  D^{1/2} ( \idm -   u_+ u_-^\top -   u_- u_+^\top ) D^{1/2},
\EEAS
with $u_+^\top u_- =0$ and $u_+, u_-$ of norm 1. Thus
we have
\BEAS
  \idm -   u_+ u_-^\top -   u_- u_+^\top 
& = &  \idm -   u_+ u_+^\top -   u_- u_-^\top +
( u_+,  u_- ) 
\big(\!\!\! \begin{array}{cc} 1\! \!\!&\!\! -1 \\
 -1 \!\!&\!\!\! 1  \end{array}\!\!\big)
(u_+ , u_-)^\top
\\
 ( \idm -   u_+ u_-^\top -   u_- u_+^\top )^{-1}
& = &  \idm -   u_+ u_+^\top -   u_- u_-^\top +
\frac{1}{0} ( u_+,  u_- ) 
\big(\!\!\! \begin{array}{cc} 1\! \!\!&\!\! 1 \\
 1 \!\!&\!\!\! 1  \end{array}\!\!\big)
(u_+ , u_-)^\top  .
\EEAS
Thus, if $v^\top 1_n = 0$, we get:
\[v^\top A^{-1}
v = v^\top \Diag(n_+ e_+ + n_- e-) v
- (v^\top e_+)^2  - (v^\top e_-)^2,
\]
which has running-time complexity $O(n)$.

\textbf{Optimization problem.}
With a regularizer $f(x) + \frac{\lambda}{2} \| x\|^2$, we obtain the problem:
\BEAS
& & \min_{ x\in \rb^d} \frac{\lambda}{2} \| x\|^2 + f(x) + \frac{1}{2} - a^\top K x + \frac{1}{2} (Kx)^\top A (Kx) \\
& & \min_{ x\in \rb^d} \max_{y \in \rb^n}
\frac{\lambda}{2} \| x\|^2 +   f(x)
 + \frac{1}{2} - a^\top K x + y^\top K x - \frac{M}{2} \| y\|^2 -  \frac{1}{2} y^\top ( A^{-1}  -  M \idm)y,
\EEAS
with $g(y) = \frac{1}{2} y^\top ( A^{-1}  -  M \idm)y$.

\section{Additional Experimental Results}
\label{app:results}
\begin{figure}
\hspace*{-1cm}
\includegraphics[scale=.4]{plot_auc_distance_auc_clusternorm_sido_1.eps}
\includegraphics[scale=.4]{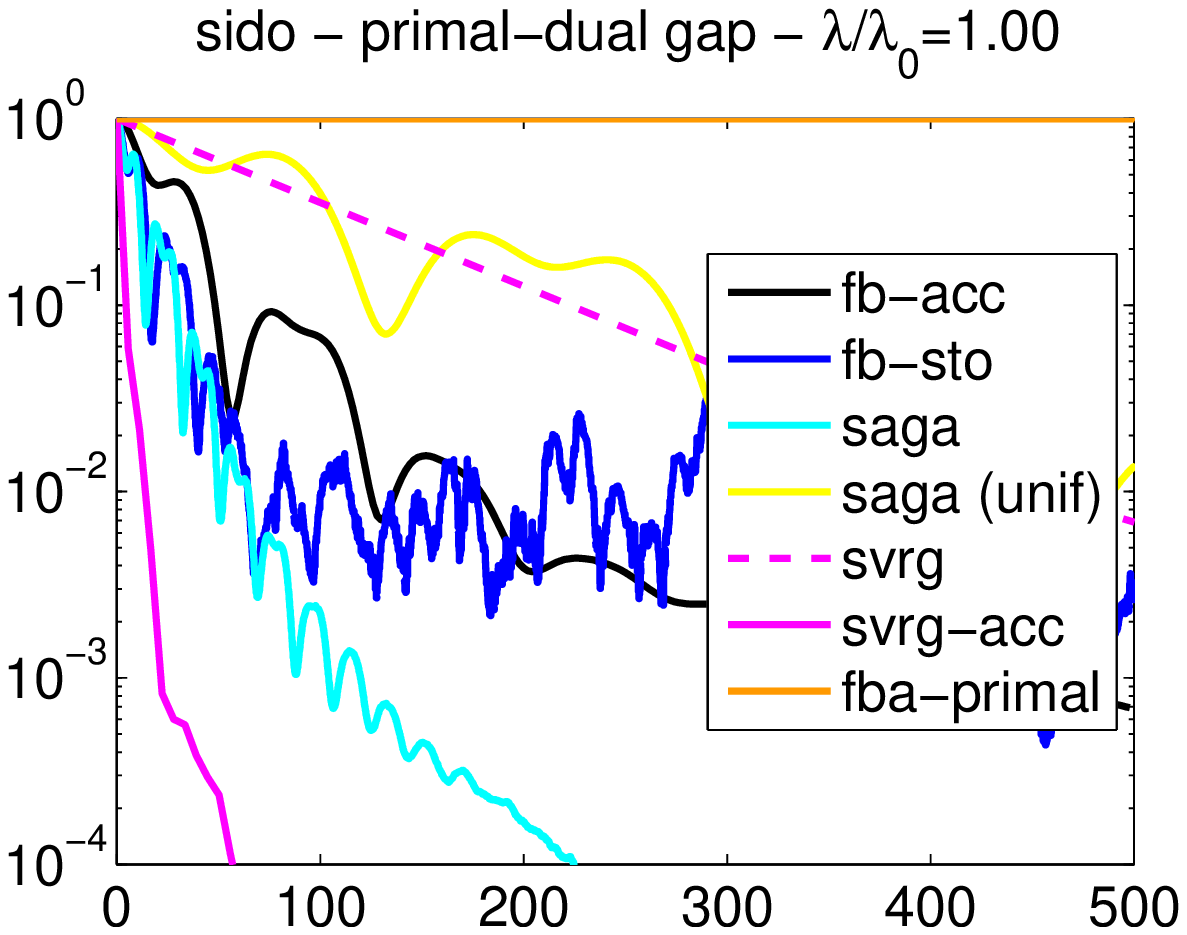}
\includegraphics[scale=.4]{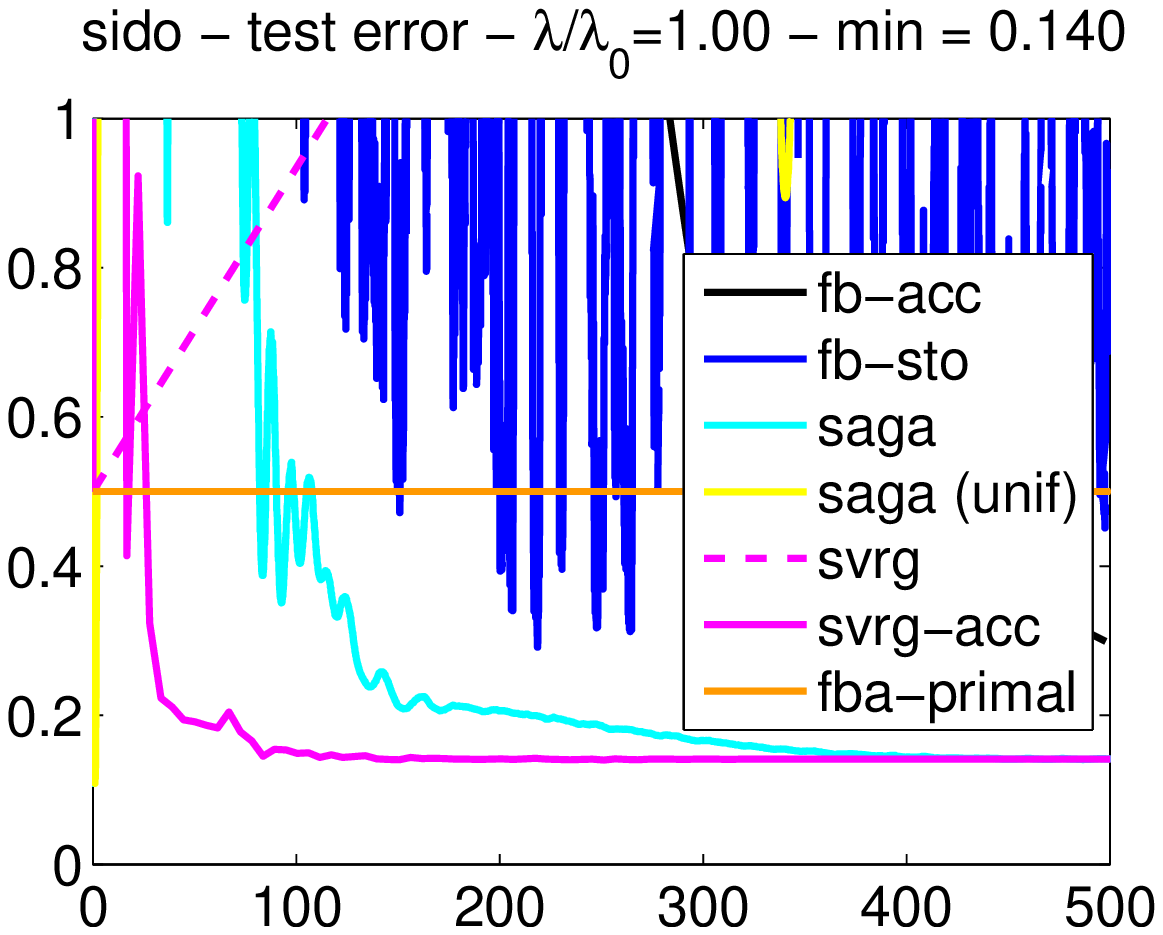}
\hspace*{-1cm}

\vspace*{1cm}

\hspace*{-1cm}
\includegraphics[scale=.4]{plot_auc_distance_auc_clusternorm_sido_1.000000e-01.eps}
\includegraphics[scale=.4]{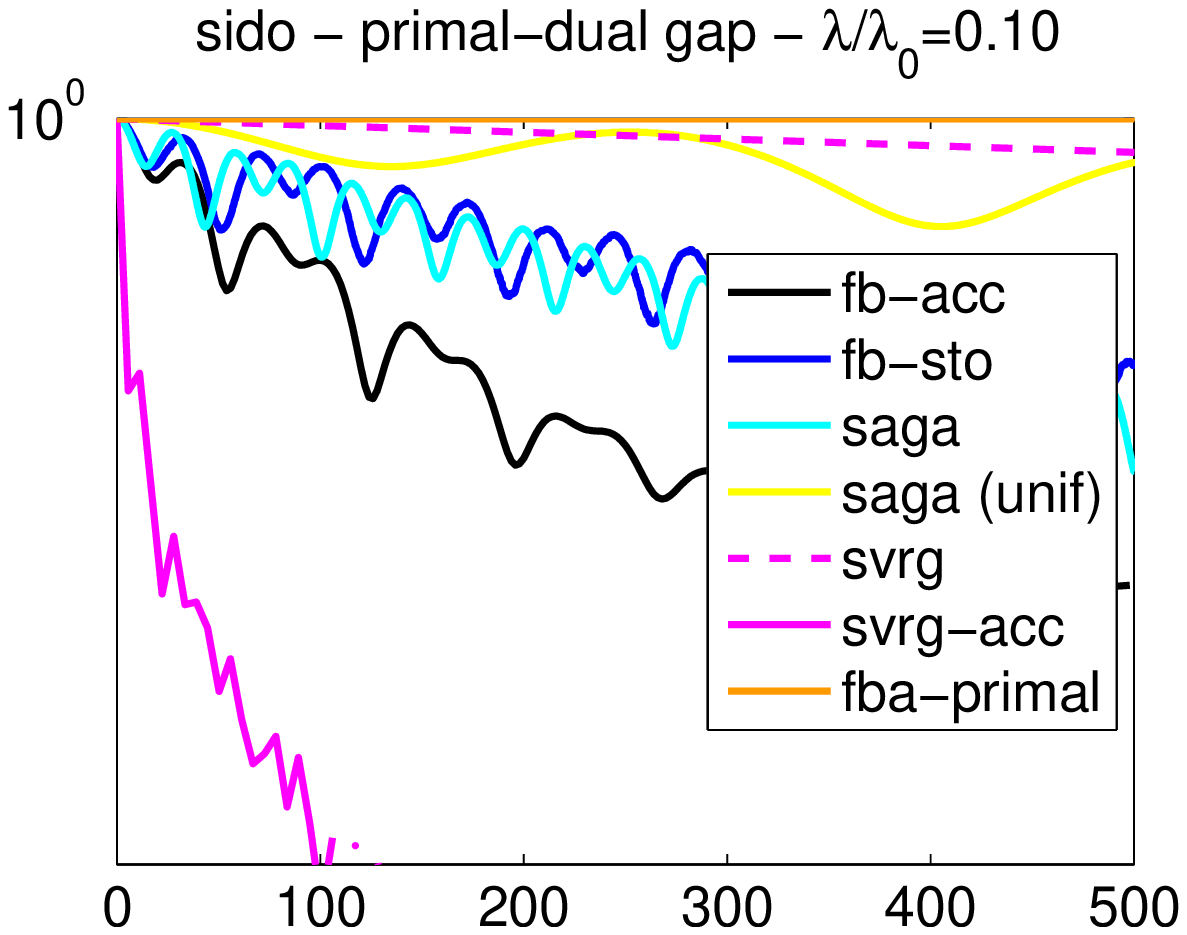}
\includegraphics[scale=.4]{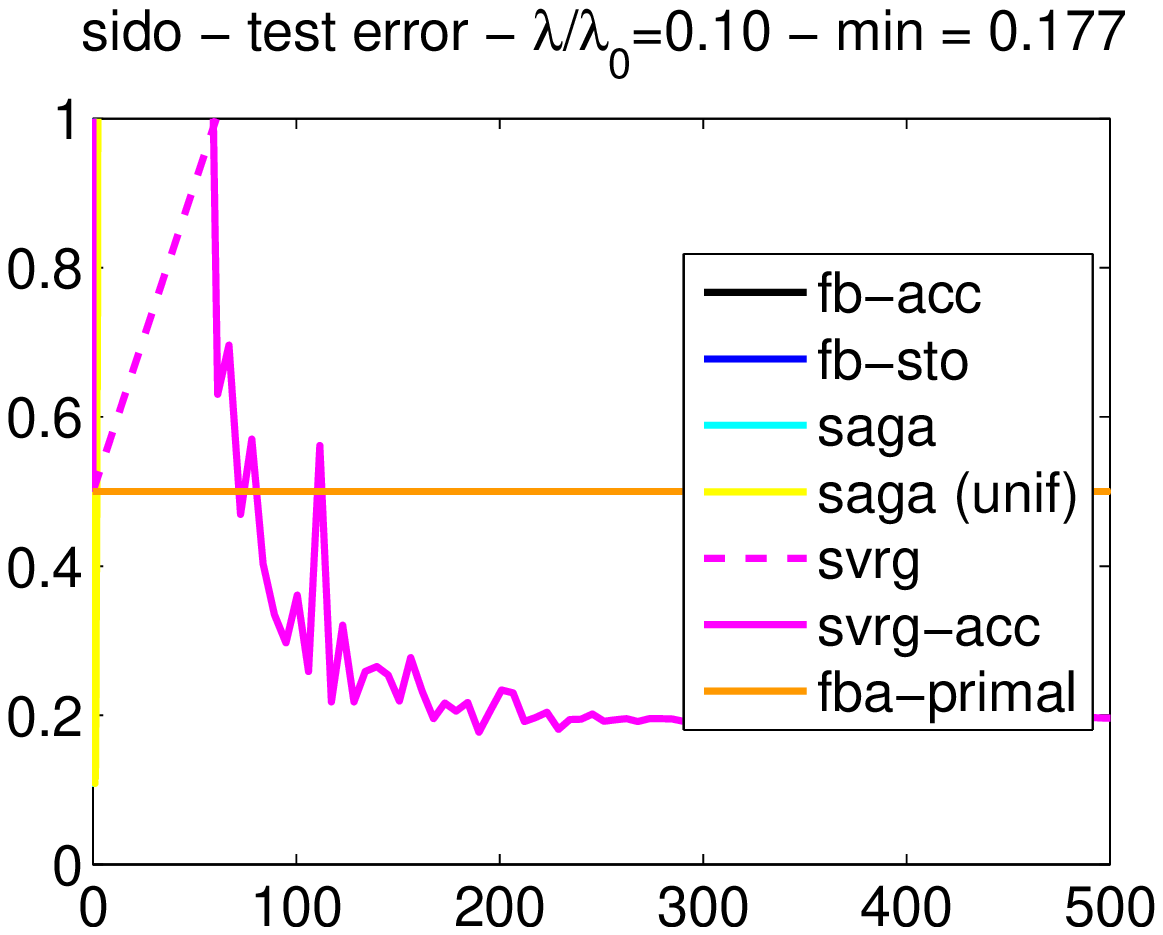}
\hspace*{-1cm}

\caption{\texttt{sido} dataset. Top: $\lambda = \lambda_0 =  \| K \|_F^2 / n^2$, Bottom: $\lambda = \lambda_0 / 10 = \frac{1}{10} \| K \|_F^2 / n^2$. AUC loss and cluster-norm regularizer. Distances to optimum, primal-dual gaps and test losses, as a function of the number of passes on the data. Note that the primal SAGA (with non-uniform sampling) cannot be used because the loss is not separable. Best seen in color.}
\label{fig:sido-auc}

\end{figure}

\begin{figure}
\hspace*{-1cm}
\includegraphics[scale=.4]{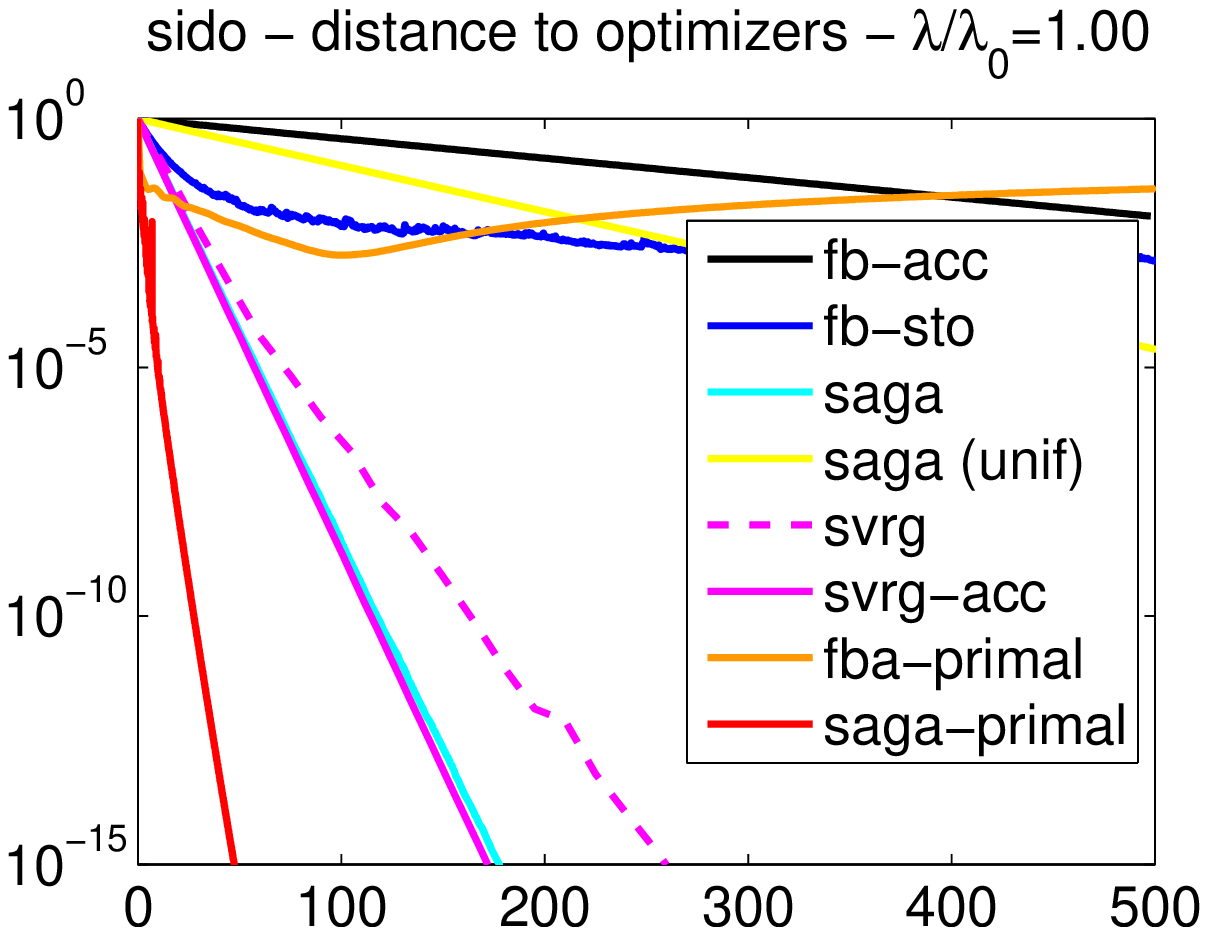}
\includegraphics[scale=.4]{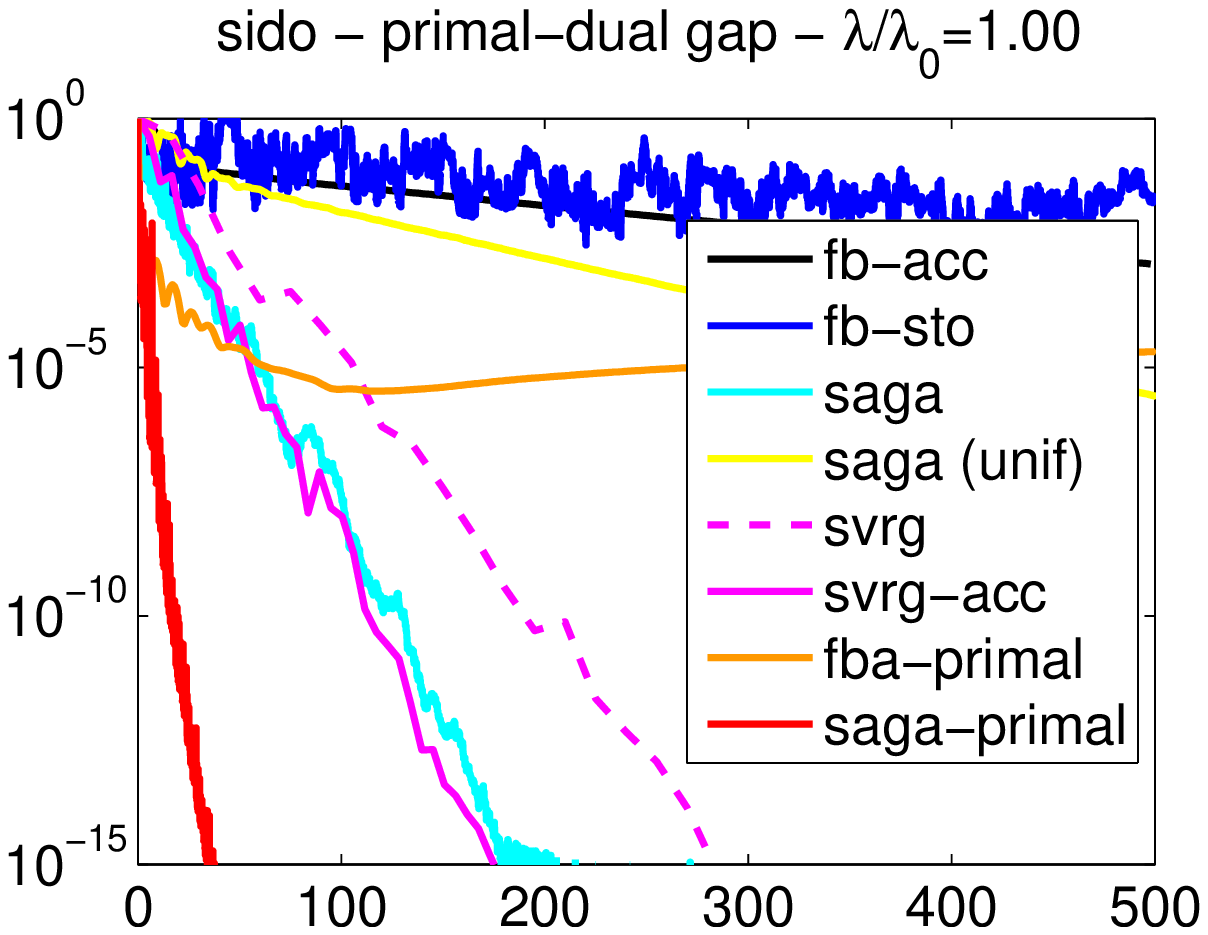}
\includegraphics[scale=.4]{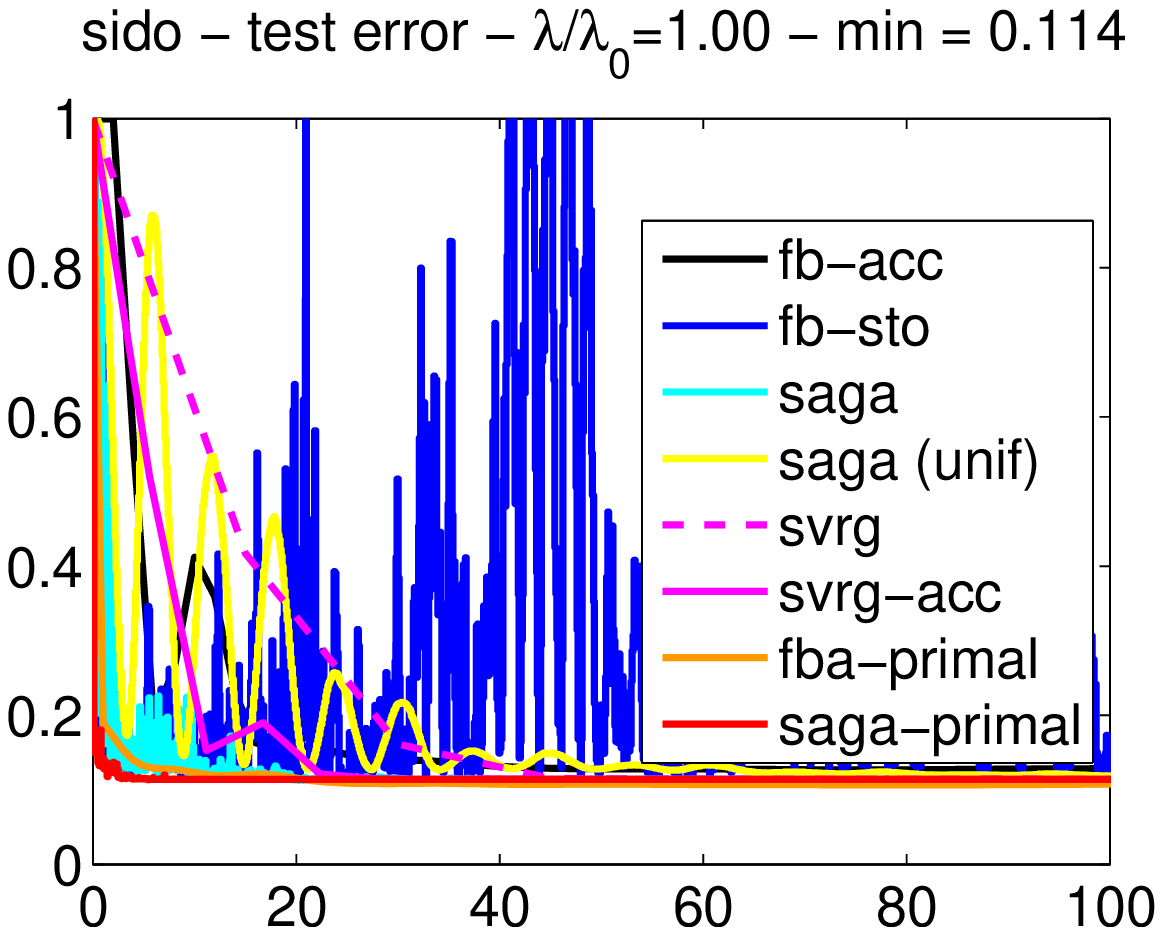}
\hspace*{-1cm}

\vspace*{1cm}

\hspace*{-1cm}
\includegraphics[scale=.4]{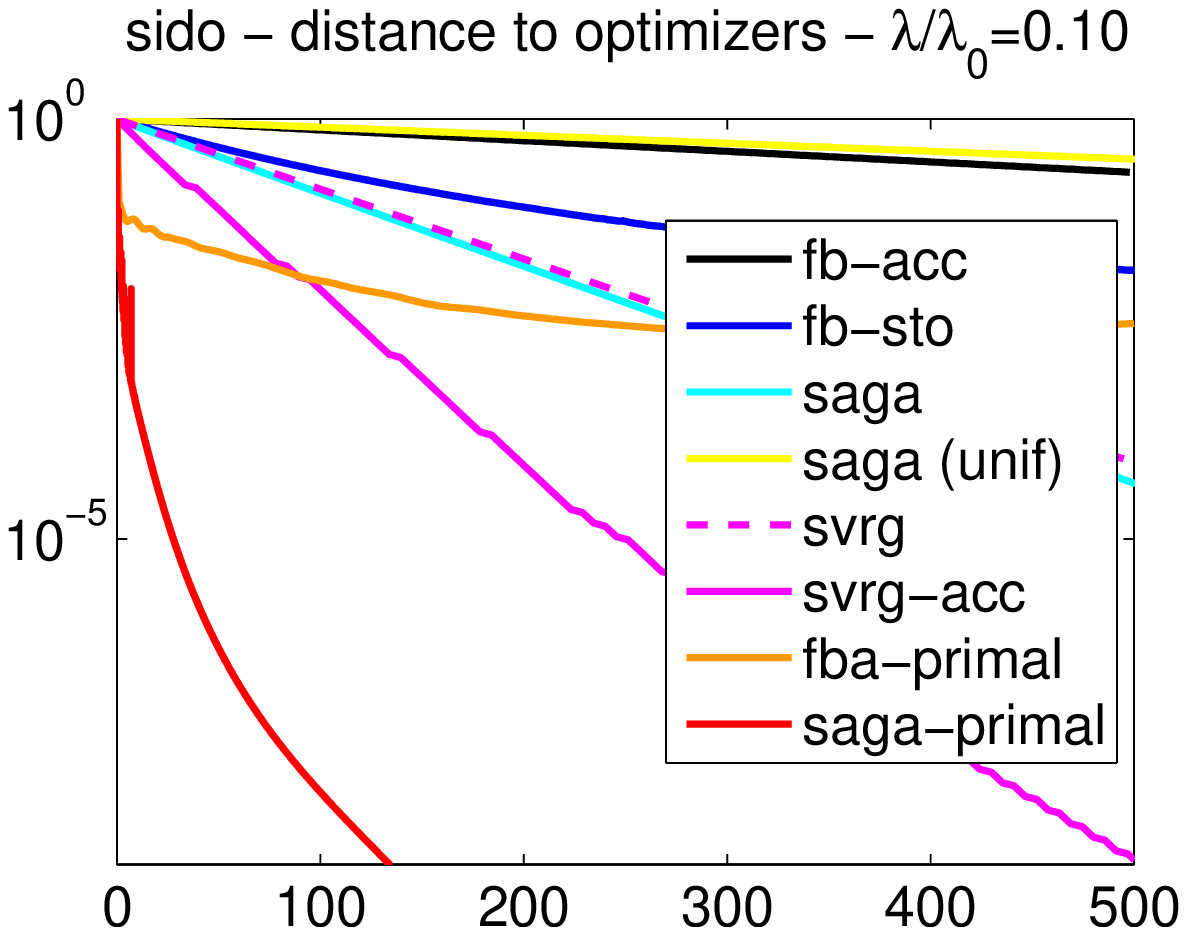}
\includegraphics[scale=.4]{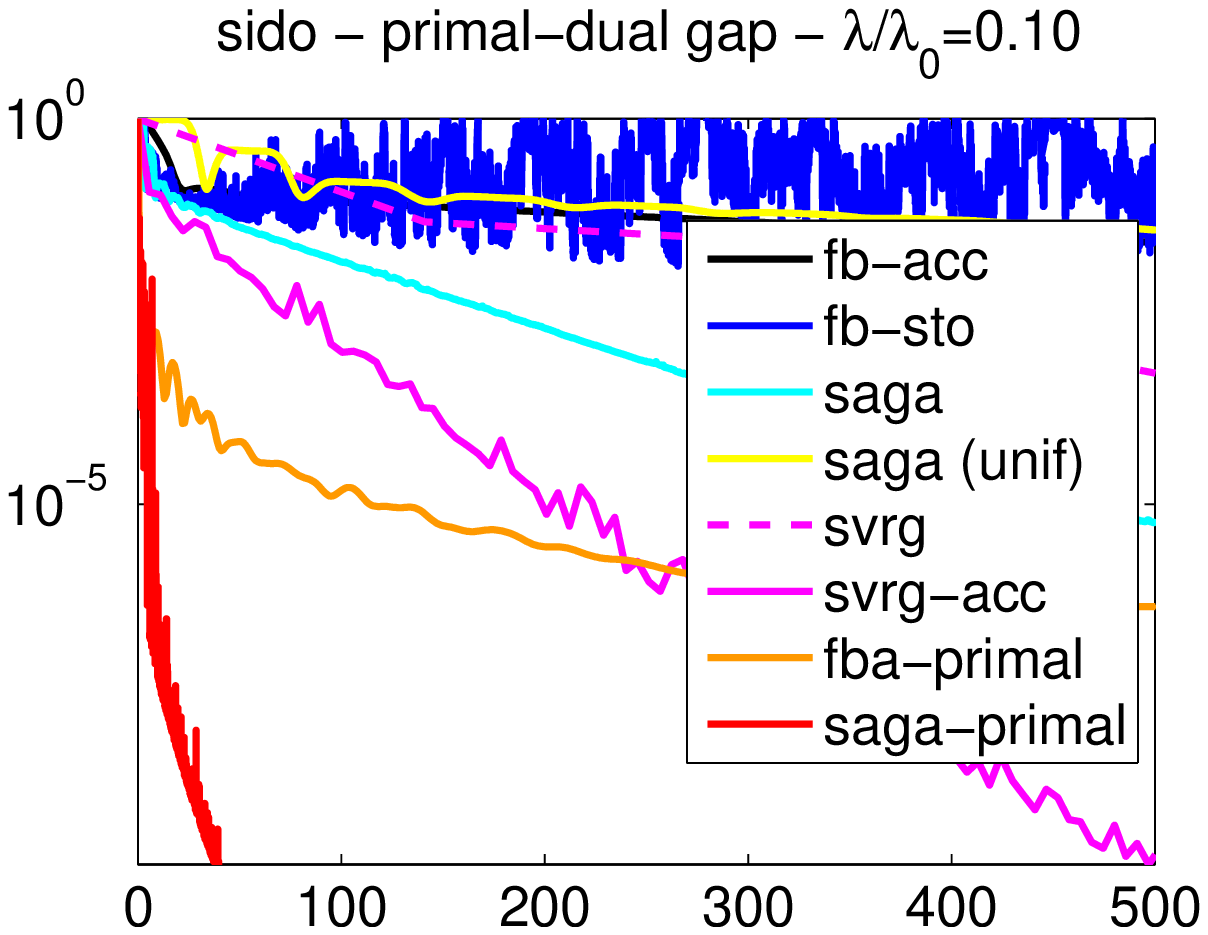}
\includegraphics[scale=.4]{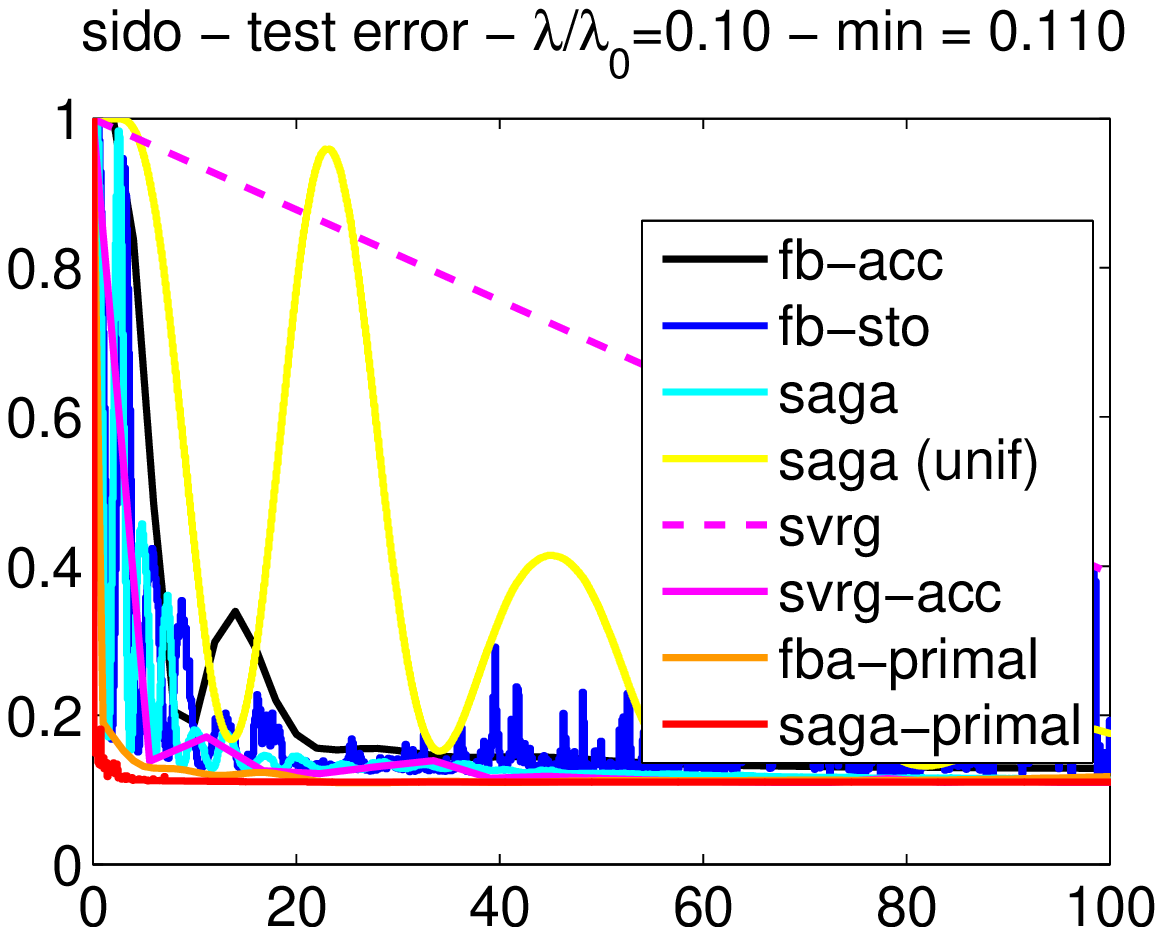}
\hspace*{-1cm}

\caption{\texttt{sido} dataset. Top: $\lambda = \lambda_0 =  \| K \|_F^2 / n^2$, Bottom: $\lambda = \lambda_0 / 10 = \frac{1}{10} \| K \|_F^2 / n^2$.  Squared loss, with $\ell_1$-regularizer. Distances to optimum, primal-dual gaps and test losses, as a function of the number of passes on the data.
Note that the primal SAGA (with non-uniform sampling) can only be used because the loss is separable. Best seen in color. }
\label{fig:sido-L2}

\end{figure}

\begin{figure}
\hspace*{-1cm}
\includegraphics[scale=.4]{plot_with_sep_distance_l1l2reg_l2loss_rcv1_1.eps}
\includegraphics[scale=.4]{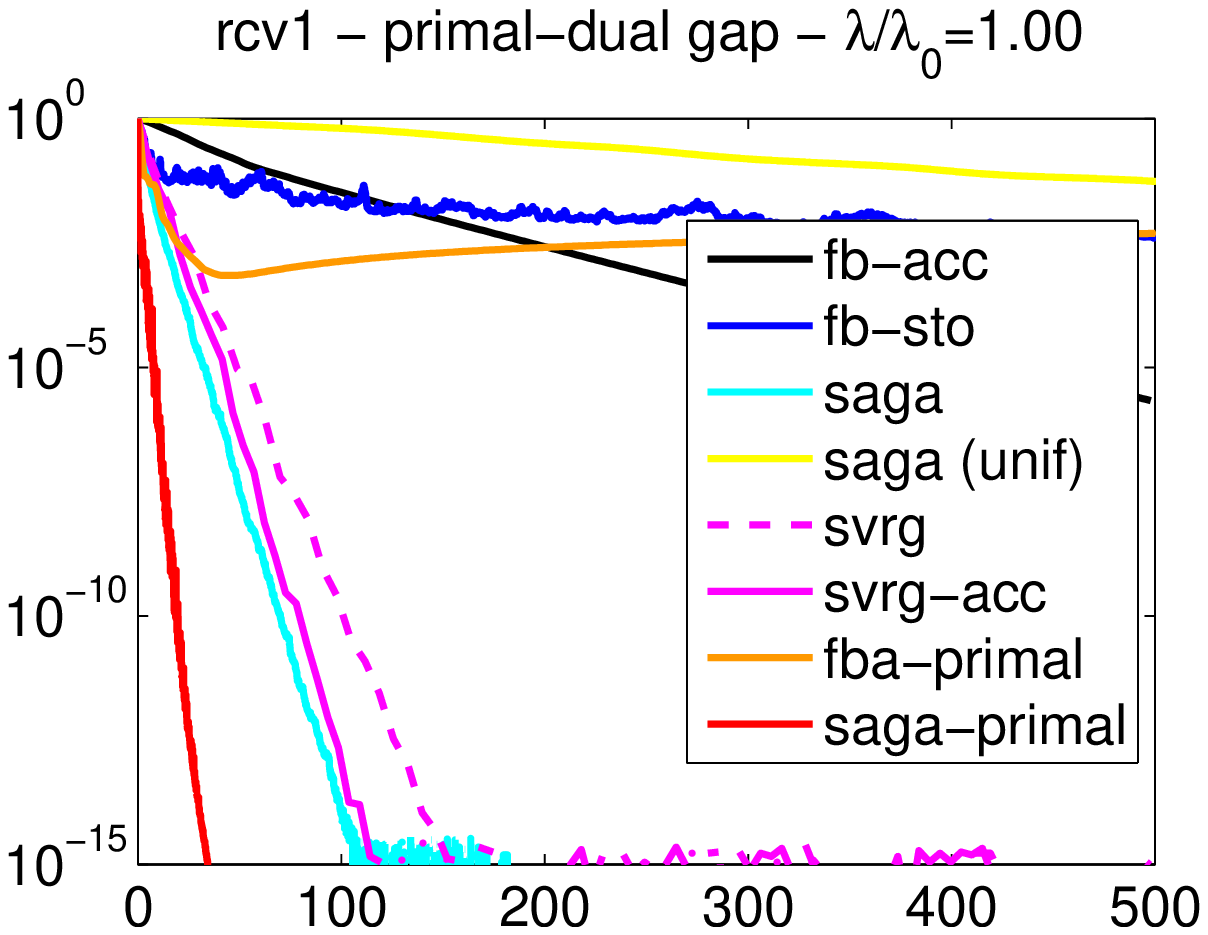}
\includegraphics[scale=.4]{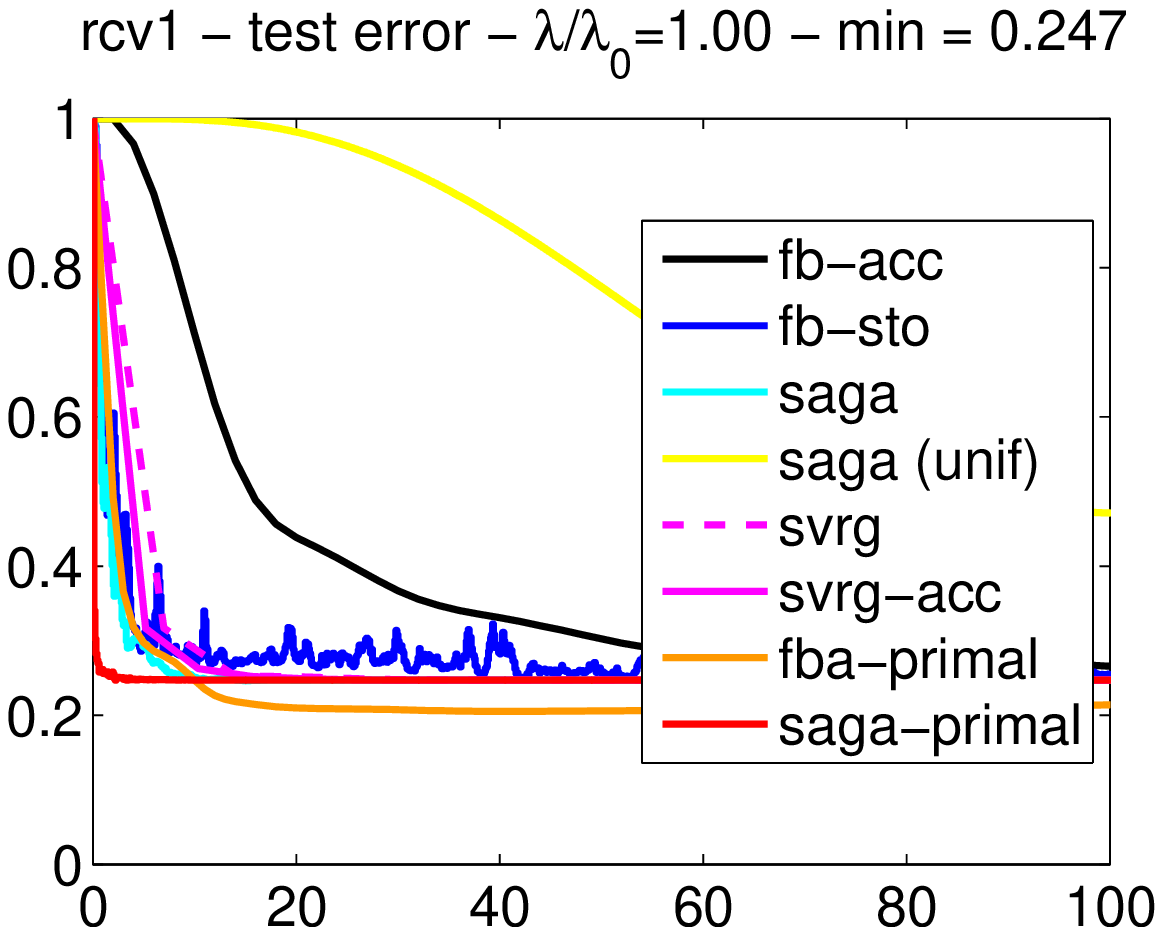}
\hspace*{-1cm}

\vspace*{1cm}

\hspace*{-1cm}
\includegraphics[scale=.4]{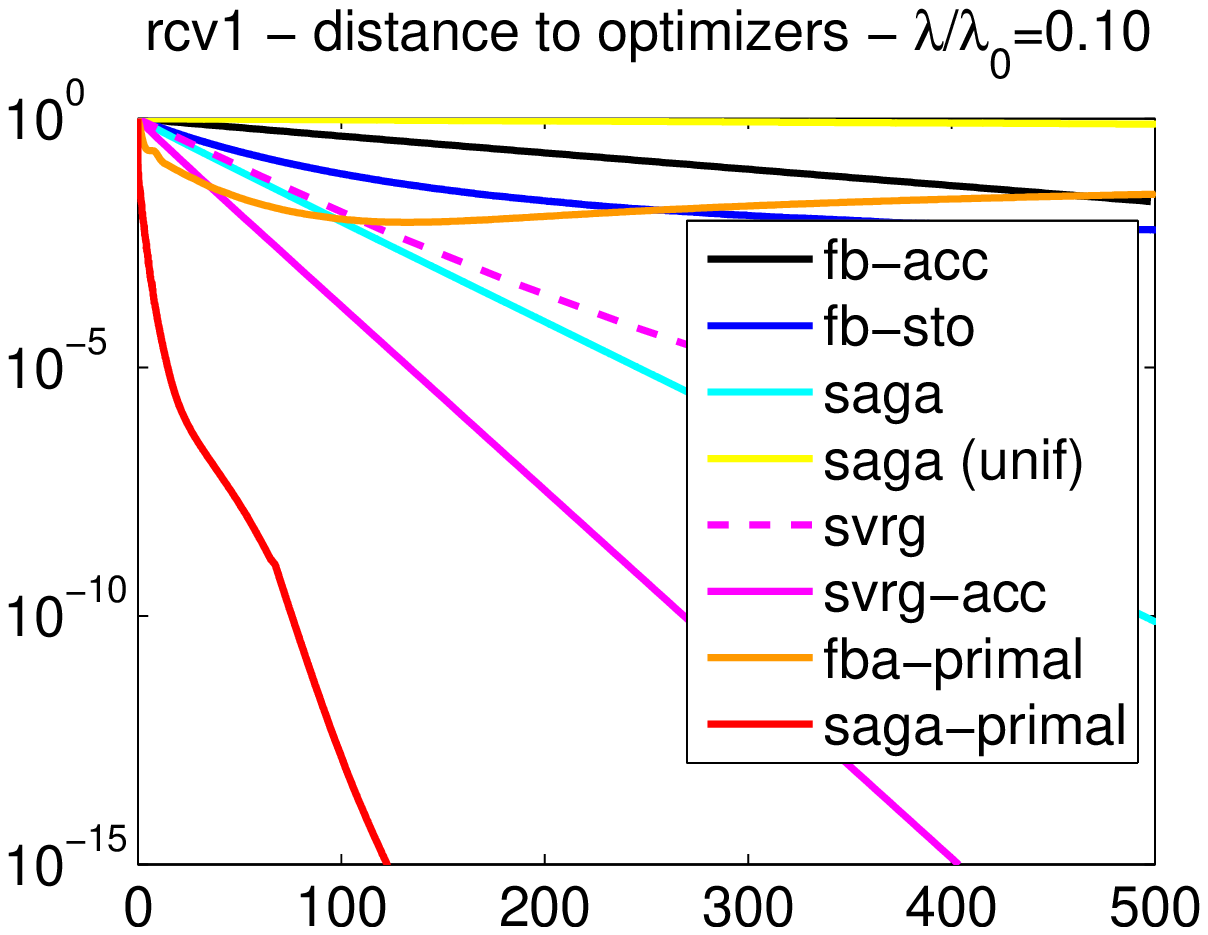}
\includegraphics[scale=.4]{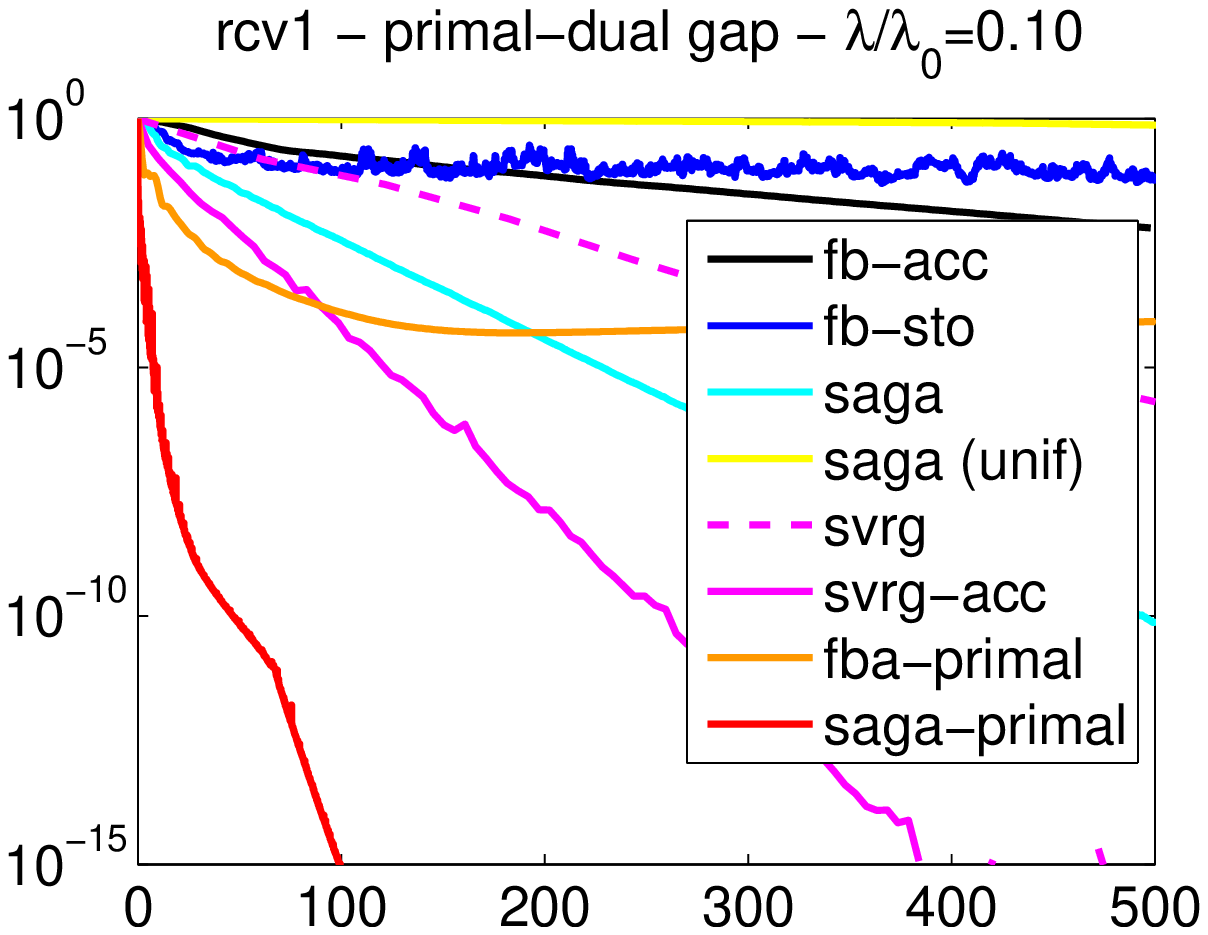}
\includegraphics[scale=.4]{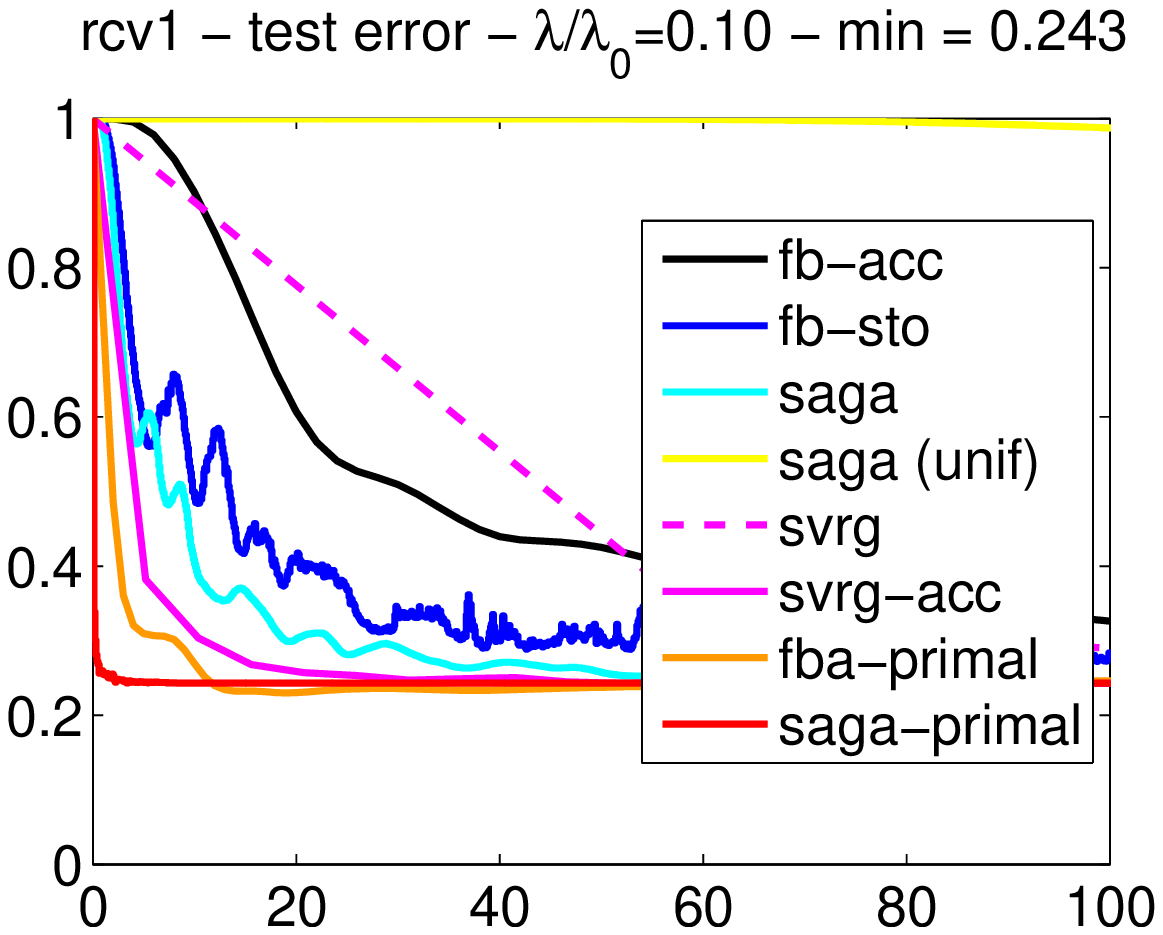}
\hspace*{-1cm}

\caption{\texttt{rcv1} dataset. Top: $\lambda = \lambda_0 =  \| K \|_F^2 / n^2$, Bottom: $\lambda = \lambda_0 / 10 = \frac{1}{10} \| K \|_F^2 / n^2$.  Squared loss, with $\ell_1$-regularizer. Distances to optimum, primal-dual gaps and test losses, as a function of the number of passes on the data. Note that the primal SAGA (with non-uniform sampling) can only be used because the loss is separable. Best seen in color.
\label{fig:rcv1-L2}
}

\end{figure}

We complement the results of the main paper in several ways: (a) by providing   all test losses, the distance to optimum $\Omega(x-x_\ast,y-y_\ast)$ in log-scale, as well as the primal-dual gaps in log-scale, as a function of the number of passes on the data. We consider the three machine learning settings:
\BIT
\item[--] Figure~\ref{fig:sido-auc}: \texttt{sido} dataset, AUC loss and cluster norm (plus squared-norm) regularizer (both non separable).
\item[--] Figure~\ref{fig:sido-L2}: \texttt{sido} dataset, square loss and $\ell_1$-norm (plus squared-norm) regularizer (both  separable).
\item[--] Figure~\ref{fig:rcv1-L2}: \texttt{rcv1} dataset, square loss and $\ell_1$-norm (plus squared-norm) regularizer (both  separable).

\EIT
We consider the following methods in all cases (all methods are run with the step-sizes proposed in their respective convergence analysis):
\BIT
\item[--] fb-acc: accelerated forward-backward saddle-point method from \mysec{batch},
\item[--] fb-sto: stochastic forward-backward saddle-point method from \mysec{sto},
\item[--] saga: our new algorithm from \mysec{saga}, with non-uniform sampling, and sampling of a single row and column per iteration,
\item[--] saga (unif): our new algorithm from \mysec{saga}, with uniform sampling, and sampling of a single row and column per iteration,

\item[--] svrg: our new algorithm from \mysec{svrg}, with non-uniform sampling, and sampling of a single row and column per iteration,

\item[--] svrg-acc: our new accelerated algorithm from \mysec{svrg}, with non-uniform sampling, and sampling of a single row and column per iteration,

\item[--] fba-primal: accelerated proximal method~\cite{Beck2009}, which can be applied to the primal version of our problem (which is the sum of a smooth term and a strongly convex term).
\EIT
Moreover, for the separable cases, we add:
\BIT
\item[--] saga-primal: SAGA with non-uniform sampling~\cite{resa}, which can only be run with separable losses.
\EIT

We can make the following observations:
\BIT
\item[--] Non-uniform sampling is key to  good performance.

\item[--] The distance to optimum (left plots) exhibits a clear linear convergence behavior (which is predicted by our analysis), which is not the case for the primal-dual gap, which does converge, but more erratically. It would be interesting to provide bounds for these as well.

\item[--] When $\lambda$ decreases (bottom plots, more ill-conditioned problems), the gains of accelerated methods with respect to non-accelerated ones are unsurprisingly larger. Note that for two out of three settings, the final test loss is smaller for the smaller regularization, and non-accelerated methods need more passes on the data to reach good testing losses.

\item[--] Primal methods  which are not using separability (here ``fba-primal'') can be run on all instances, but are not competitive. Note that in some situations, they achieve early on good performances (e.g., Figure~\ref{fig:sido-L2}), before getting caught up by stochastic-variance-reduced saddle-point techniques (note also that since these are not primal-dual methods, we compute dual candidates through the gradient of the smooth loss functions, which is potentially disadvantageous).

\item[--] Primal methods that use separability  (here ``saga-primal'')  cannot be run on non-separable problems, but when they can run, they are still significantly faster than our saddle-point techniques. We believe that this is partly due to adaptivity to strong convexity (the convergence bounds for the two sets of techniques are the same for this problem).

\EIT

\end{document}